\documentclass[lettersize,journal]{IEEEtran}
\usepackage{amsmath,amsfonts}
\usepackage{array}
\usepackage{textcomp}
\usepackage{stfloats}
\usepackage{url}
\usepackage{verbatim}
\usepackage{graphicx}
\usepackage{cite}
\usepackage{subfigure}
\usepackage{caption}
\usepackage{tabularx}
\usepackage{multirow}
\usepackage{makecell}
\usepackage{booktabs}
\usepackage{array} 
\usepackage{setspace}
\usepackage{ulem}
\usepackage{amsmath}
\usepackage[linesnumbered, ruled]{algorithm2e}
\usepackage{hyperref}
\SetKwRepeat{Do}{do}{while}%

\begin{document}

\title{SMPISD-MTPNet: Scene Semantic Prior-Assisted Infrared Ship Detection Using Multi-Task Perception Networks}
\author{Chen Hu, Xiaogang Dong, Yian Huang,~\IEEEmembership{Student Member,~IEEE,} Lele Wang, Liang Xu, Tian Pu, and Zhenming Peng,~\IEEEmembership{Member,~IEEE}

\thanks{Manuscript received XXX XXX, XXX; revised XXX XXX, XXX.}
\thanks{This work was supported by the Natural Science Foundation of Sichuan Province of China (Grant No.2022NSFSC40574 and Grant No.2022YFG0178) and partially supported by the National Natural Science Foundation of China (Grant No.61775030 and Grant No.61571096). \textit{(Corresponding authors: Zhenming Peng; Tian Pu.)}}
\thanks{The authors are with the School of Information and Communication Engineering and the Laboratory of Imaging Detection and Intelligent Perception, University of Electronic Science and Technology of China, Chengdu 610054, China (e-mail: 202221011506@std.uestc.edu.cn; dongxg1516@126.com; huangyian@std.uestc.edu.cn; 202321011802std.uestc.edu.cn; nyaluderlu@gmail.com; putian@uestc.edu.cn; zmpeng@uestc.edu.cn).}
}

\markboth{Journal of \LaTeX\ Class Files,~Vol.~XXX, No.~XXX, XXX~XXX}%
{Shell \MakeLowercase{\textit{et al.}}: A Sample Article Using IEEEtran.cls for IEEE Journals}

\IEEEpubidadjcol 

\maketitle

\begin{abstract}
Infrared ship detection (IRSD) has received increasing attention in recent years due to the robustness of infrared images to adverse weather. However, a large number of false alarms may occur in complex scenes. To address these challenges, we propose the Scene Semantic Prior-Assisted Multi-Task Perception Network (SMPISD-MTPNet), which includes three stages: scene semantic extraction, deep feature extraction, and prediction. In the scene semantic extraction stage, we employ a Scene Semantic Extractor (SSE) to guide the network by the features extracted based on expert knowledge. In the deep feature extraction stage, a backbone network is employed to extract deep features. These features are subsequently integrated by a fusion network, enhancing the detection capabilities across targets of varying sizes. In the prediction stage, we utilize the Multi-Task Perception Module, which includes the Gradient-based Module and the Scene Segmentation Module, enabling precise detection of small and dim targets within complex scenes. For the training process, we introduce the Soft Fine-tuning training strategy to suppress the distortion caused by data augmentation. Besides, due to the lack of a publicly available dataset labelled for scenes, we introduce the Infrared Ship Dataset with Scene Segmentation (IRSDSS). Finally, we evaluate the network and compare it with state-of-the-art (SOTA) methods, indicating that SMPISD-MTPNet outperforms existing approaches. The source code and dataset for this research can be accessed at https://github.com/greekinRoma/KMNDNet.
\end{abstract}


\section{Introduction}
\IEEEPARstart{S}{HIP} detection is vital for many civilian and military tasks, such as marine resource management, marine search and rescue, and coastal monitoring. However, the complexity and diversity of the scene result in low accuracy and poor robustness of infrared ship detection (IRSD). For example, clouds obscure parts of the ships, preventing the targets from being detected. Therefore, IRSD is a valuable and challenging research topic.

Ship detection mainly relies on three technologies: synthetic aperture radar (SAR) \cite{Zhang1}, visible remote sensing (VRS), and infrared imaging. Compared with the other two technologies, infrared imaging provides good concealment and is less susceptible to changes in lighting. Besides, the quality of infrared images has improved with advancements in infrared technology. Therefore, conducting research on ship detection based on infrared images is highly valuable.

IRSD encounters multiple challenges: 1) In satellite remote sensing, long-range infrared imaging typically produces low-resolution images of ships, resulting in the lack of apparent characteristics of the ship and the inability to show the detailed features of the ship. 2) There are several complex scenes. Nearshore dams and buildings resembling ships lead to false alarms. Additionally, densely distributed ships can result in missed detections. Offshore, the presence of reefs, clouds, and long ship tails can obstruct IRSD. 3) The wide variety of ship types and their substantial differences present significant challenges to the modelling. 4) No infrared ship datasets are available for scene perception, highlighting the need to create a specific dataset for this research.

Current detection can be divided into two main categories: model-driven and data-driven methods. Model-driven methods rely on expert knowledge. There are many contributions of researchers about this. For example, many researchers depend on the sparse features of targets and apply low-rank sparse decomposition techniques for the detection of small targets \cite{6}, \cite{7}, \cite{8}, \cite{9}, \cite{10}. Zhu et al. \cite{11} use texture and shape features to detect the ship. However, these methods cannot cope with complex backgrounds \cite{12}. Unlike model-driven methods, data-driven methods, especially neural networks, have a better capability for complex backgrounds. Much research currently employs deep networks for ship detection in SAR images \cite{DEFORM-FPN,MASK-ATTENTION}. Most neural networks for detection are based on convolutional networks \cite{18} or Transformer \cite{19}. With the development of deep learning, many excellent detection networks have been proposed, such as Faster Region-based Convolutional Neural Network (Faster RCNN)\cite{20}, You Only Look Once (YOLO) \cite{21}, Single Shot Multibox Detector (SSD) \cite{22} and Swin-Transformer \cite{23}. These networks mainly rely on high-level semantics to distinguish between targets and false alarms \cite{24}. However, as mentioned above, infrared ships lack the apparent characteristics. So, off-the-shelf detection networks are inappropriate for IRSD. In response, some researchers are exploring interpretable networks \cite{13,EIRNet,14,15,16,17}. 

Based on the above analysis, we introduce the Scene Semantic Prior-Assisted Multi-Task Perception Network (SMPISD-MTPNet), an end-to-end detection network. Our network has three stages: scene semantic extraction, deep feature extraction, and prediction. In scene semantic extraction, we introduce the Scene Semantic Extractor (SSE), which uses local differences to identify the scene and local contrasts to recognize the candidate region. In the deep feature extraction stage, we use CSPDarkNet53 \cite{1} as the backbone and employ a simplified Feature Pyramid Network (FPN) \cite{2} to fuse the features from varied layers. In the prediction stage, we have developed the Multi-Task Perception Module, including the Gradient-based Module, which utilizes the gradient values to strengthen the features of small targets, and the Scene Segmentation Module, which predicts the targets and scenes. To address the lack of publicly available datasets with labels for scenes, we have developed a new dataset named Infrared Ship Dataset with Scene Segmentation (IRSDSS). Additionally, we have implemented a new strategy named Soft Fine-tuning to reduce the data distortion caused by data augmentation. The main contributions of this research are as follows:

\begin{itemize}
 \item[1)] To address the lack of apparent features, we introduce the SSE to enrich the semantics of each pixel and the Gradient-based Module designed to capitalize on the differences between the background and targets. 
 \item[2)] We introduce multi-task perception by incorporating scene segmentation into the heads to reduce distractions from complex scenes. This addition enables precise identification of various backgrounds and effectively suppresses false alarms in targeted scenarios.
 \item[3)] To overcome the challenge of modelling diverse target characteristics, we employ data augmentation at the input stage to boost network generalization and introduce Soft Fine-tuning, a novel training strategy to mitigate distortion from data augmentation. Additionally, integrating a simplified FPN in our framework ensures that the output feature layers effectively capture essential information across various target sizes.
 \item[4)] Due to the scarcity of datasets with scene labels, we have created a new dataset named IRSDSS. To our knowledge, this dataset is unique in that it contains scene labels.
\end{itemize}

We organize the remainder of this article as follows: Section II reviews related work in recent years; Section III details our newly created IRSDSS dataset; Section IV outlines the structure design of SMPISD-MTPNet; Section V analyzes the performance of the proposed network structure through ablation and comparative experiments; and finally, Section VI concludes the research findings.

\section{Related Work}

\subsection{Model-driven Methods}

Model-driven methods for ship detection mainly include three steps: land-sea segmentation, candidate region identification, and target confirmation. 

Land-sea segmentation, which separates the ocean region from the land region in the land-sea background image, can reduce the impact of complex land on ship detection, thus improving the accuracy and influence of inshore ship detection. This step mainly relies on the difference between ocean and land imaging features for image segmentation. Wang et al. \cite{25} use the difference in texture between the ocean and land \cite{26} for sea-land segmentation. Zha et al. \cite{27} enhance the performance of simple linear iterative clustering \cite{28} by incorporating colour and texture features, resulting in improved outcomes. In addition to land-sea segmentation, Corbane et al. \cite{29} conduct cloud-sea segmentation to reduce the false alarms caused by clouds. 

Candidate region identification in target detection involves identifying all regions potentially containing targets, encompassing both actual targets and false alarms. Zhu et al. \cite{11} modelled ship edge features and extracted corresponding regions. Recently, some approaches based on the human vision system (HVS) have been proposed and are used for ship detection \cite{6}, \cite{30}, \cite{31}, \cite{32}, \cite{33}, \cite{34}. Hou et al. \cite{35} use the neighbourhood average deviation (NAD) to detect regions of interest (ROIs). Wang et al. \cite{36} use the local contrast measure (LCM) for target extraction.

Target confirmation relies on the features of ROIs and eliminates false alarms. This step often consists of two stages: feature extraction and classifier. In the feature extraction stage, we use the feature extraction algorithms to get the features of the candidate regions for further classification. For example, Zhu et al. \cite{11} extract the texture and shape features. In the classifier stage, we often use the features obtained by the last stage to define which regions are the targets. For instance, Xia et al. \cite{37} used a support vector machine (SVM) to confirm ship targets.

Model-driven methods are based on expert knowledge and are well-interpretable. However, they are not able to cope with complex backgrounds.

\subsection{Data-driven Methods}

With the development of computers, more attention is being paid to data-driven methods. These methods are divided into two main categories: convolution neural network-based (CNN-based) and transformer-based networks.

CNN-based networks consist of two main categories: one-stage and two-stage detection networks. Many well-known one-stage networks exist, like Faster R-CNN \cite{20}. Besides, the one-stage networks like SSD \cite{22} and YOLO \cite{21} perform well. Based on generic detection networks, many researchers modify these networks to fit the characteristics of various targets. For instance, Chen et al. \cite{38} refined YOLOV3 \cite{39} by incorporating lightweight dilated convolution modules, enhancing its computational efficiency.

Transformers \cite{19}, originally developed for Natural Language Processing (NLP), rely on self-attention mechanisms to process data. After the Vision Transformer(ViT) is introduced to the detection tasks \cite{40}, people gradually use the structure of ViT to detect the target, like the DEtection TRansformer (DETR) \cite{41}. However, transformers-based networks focus more on global semantics, sometimes overshadowing the local features critical for accurate ship detection.
 
Additionally, some researchers have developed these architectures that combine the strengths of CNNs and transformers. They utilise CNNs for their ability to capture local details and transformers for their superior handling of global contexts, thereby improving ship detection accuracy \cite{42}.

\begin{figure}
 \centering
 \subfigure[]
 {
 \includegraphics[scale=.15]{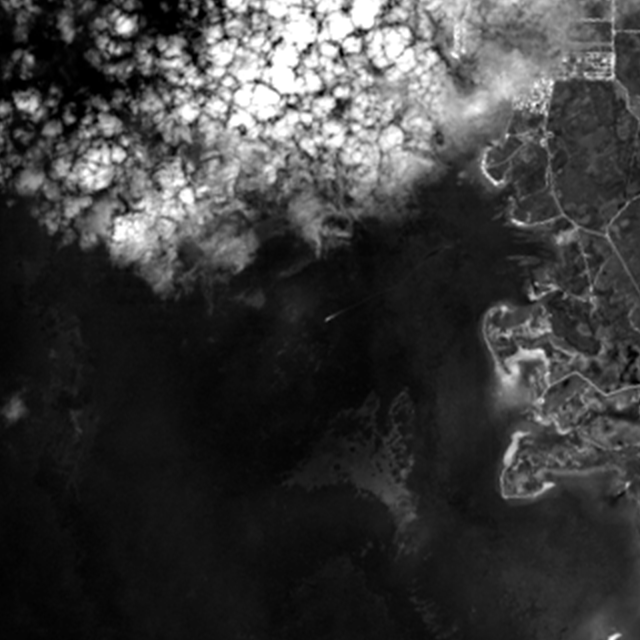}
 }
 \subfigure[]
 {
 \begin{minipage}[b]{.4\linewidth}
 \centering
 \includegraphics[scale=.15]{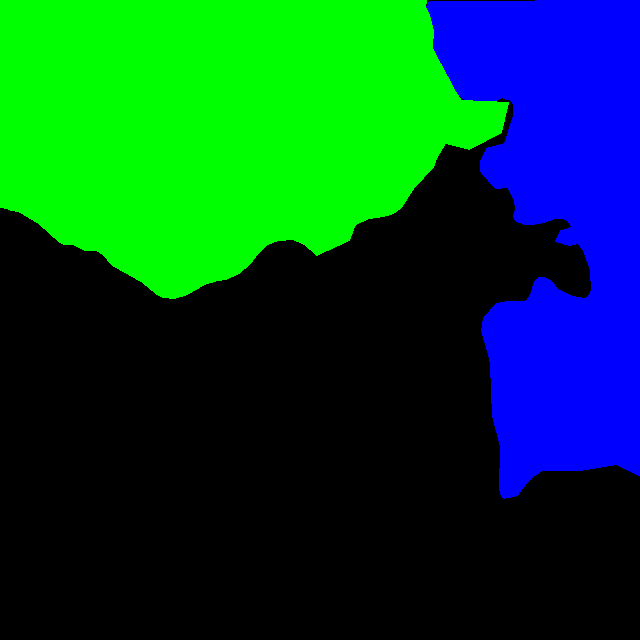}
 \end{minipage}
 }
 \caption{Infrared image and corresponding mask image: (a) Infrared image; (b) In the mask image corresponding to the infrared image, blue indicates land masks and green denotes cloud masks.}
 \label{fig_12}
\end{figure}

\begin{figure}
 \centering
 \subfigure[]
{
 \begin{minipage}[b]{.29\linewidth}
 \centering
 \includegraphics[scale=.2]{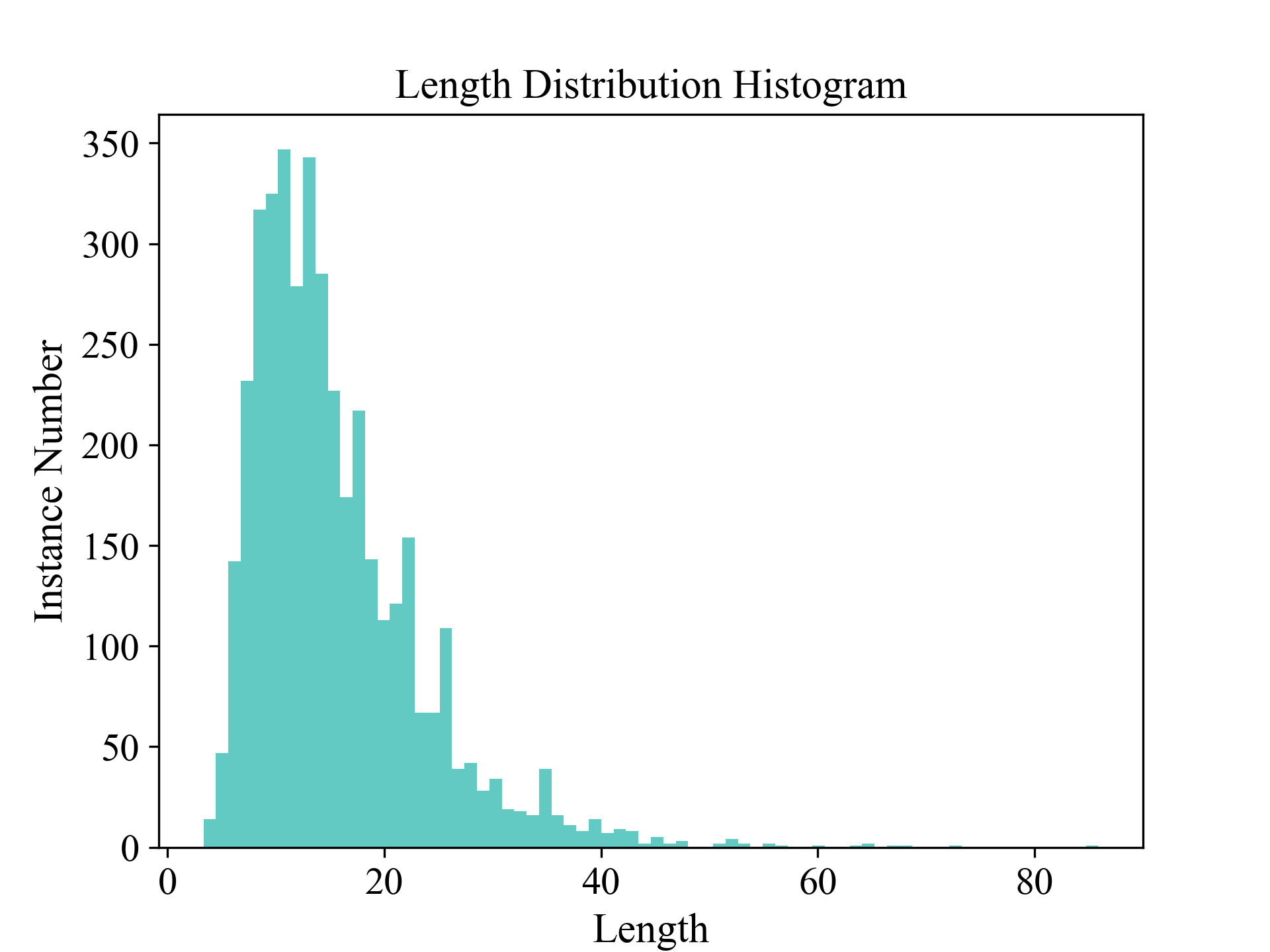}
 \end{minipage}
}
\subfigure[]
{
 	\begin{minipage}[b]{.29\linewidth}
 \centering
 \includegraphics[scale=.2]{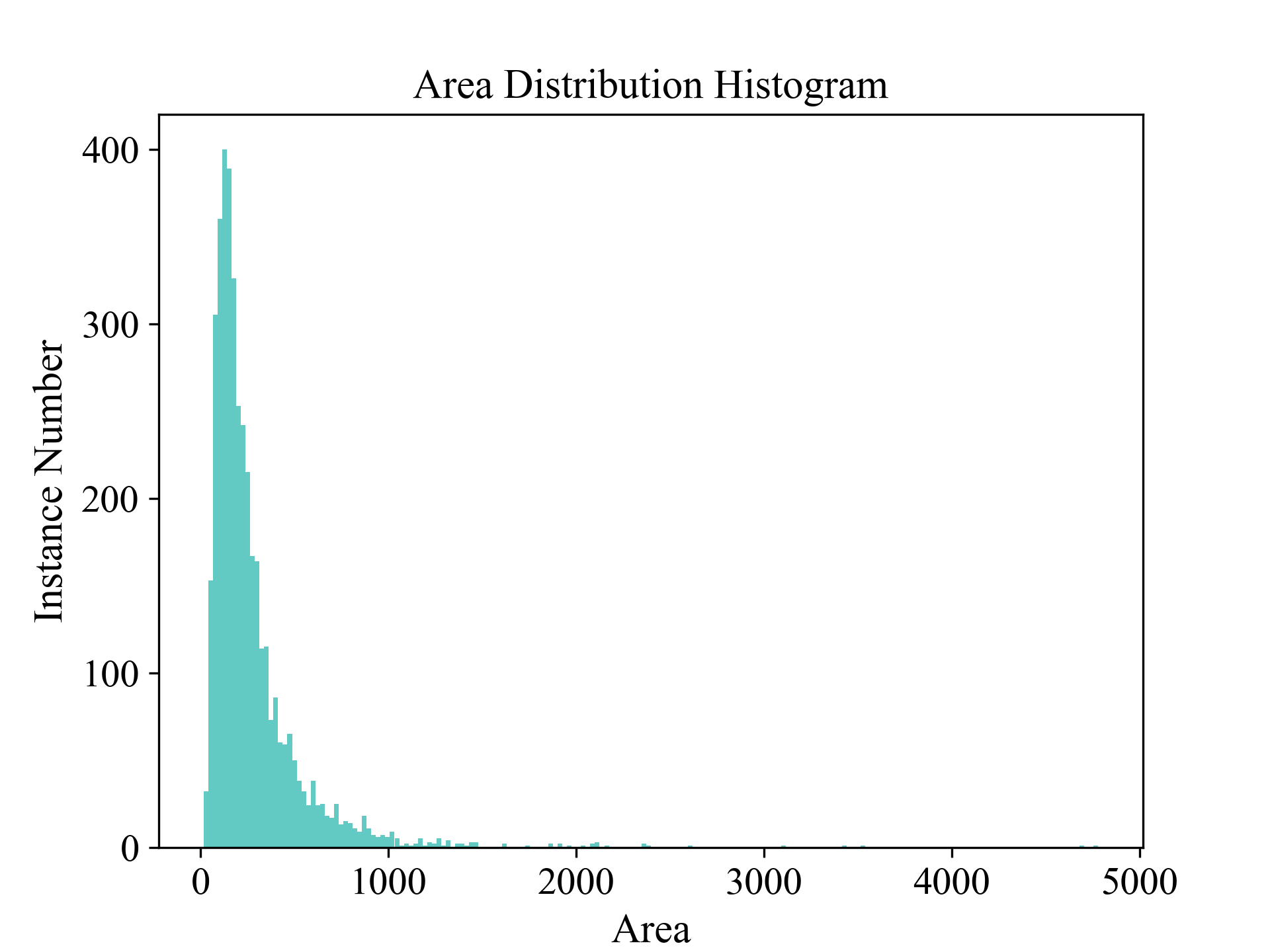}
 \end{minipage}
}
\subfigure[]
{
 	\begin{minipage}[b]{.29\linewidth}
 \centering
 \includegraphics[scale=.2]{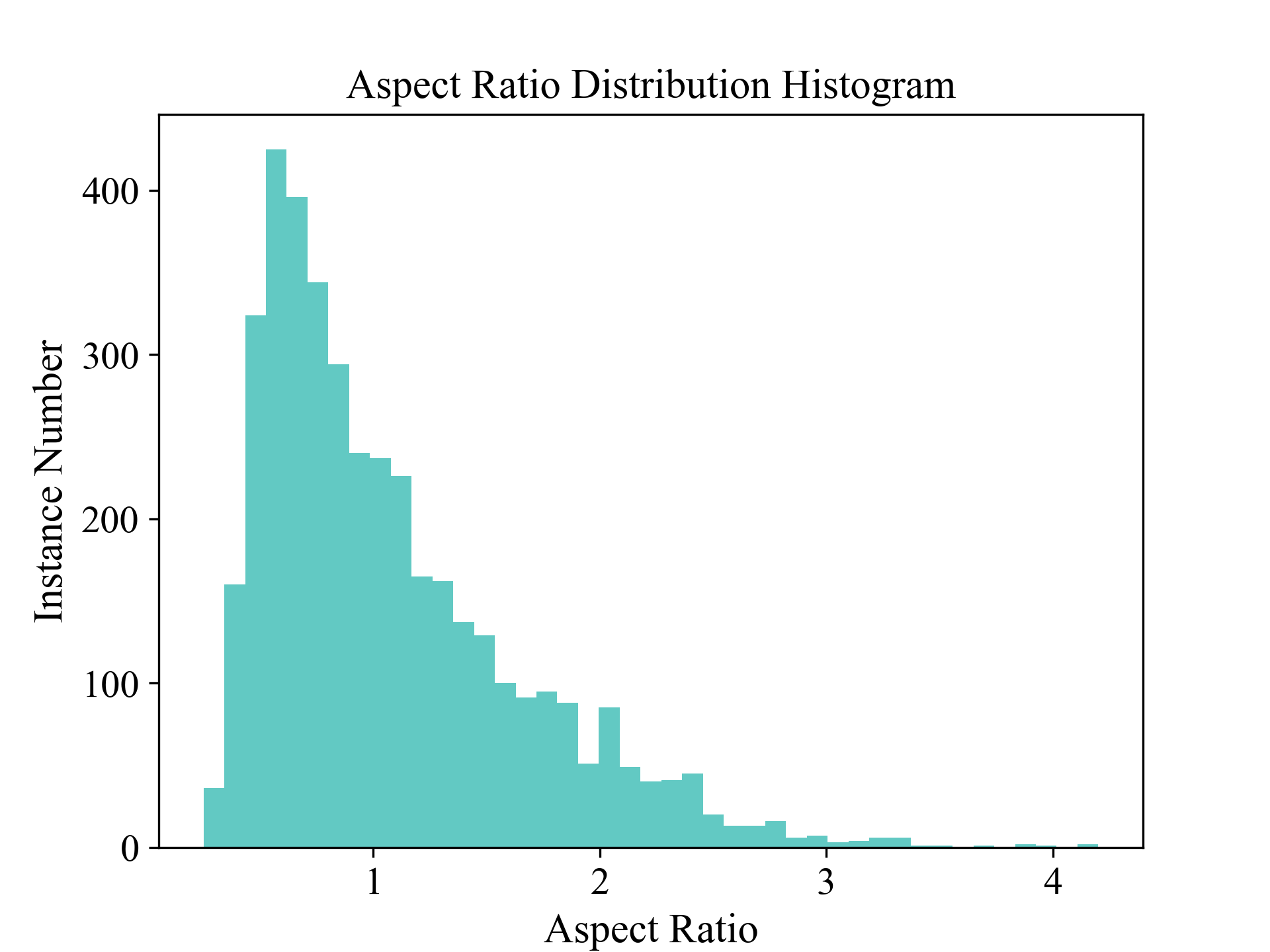}
 \end{minipage}
}

\caption{Statistics of instances in IRSDSS: (a) Instance Area, (b) Height, and (c) Aspect Ratio.}
\vspace{-.5cm}
\label{comparsion_2}
\end{figure}
\subsection{Dual-driven Methods}
Dual-driven methods enhance neural network performance by combining expert knowledge with neural networks. Chen et al. \cite{30} introduce local contrast into neural networks. Han et al. \cite{13} improve the precision of their methods by employing convolutional networks to predict manual features in candidate regions, effectively using these features to reduce false alarms. Wang et al. \cite{44} developed a cascaded decision framework that integrates global spectral feature learning via Fourier transform with local feature learning from a lightweight classification network, enhancing the analytical capabilities of the network. SVDNet based on singular value decomposition \cite{45} optimizes network efficiency by implementing singular value decomposition on the model matrix. Wu et al. \cite{15} simulated robust principal component analysis (RPCA) \cite{46} by convolution for small target detection. In the research, we optimize the performance of data-driven neural networks by introducing model-driven modifications.

\begin{figure}[htbp]
\centering
\subfigure[]
{
 \begin{minipage}[b]{.21\linewidth}
 \centering
 \includegraphics[scale=0.085]{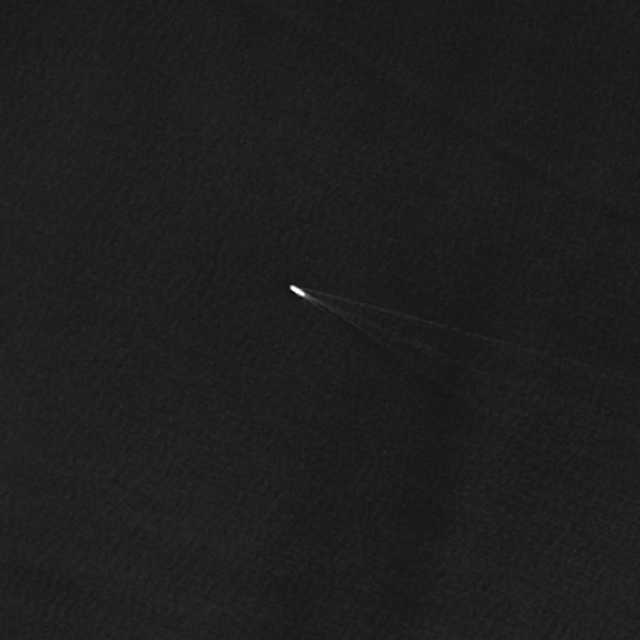}
 \end{minipage}
 \begin{minipage}[b]{.21\linewidth}
 \centering
 \includegraphics[scale=0.085]{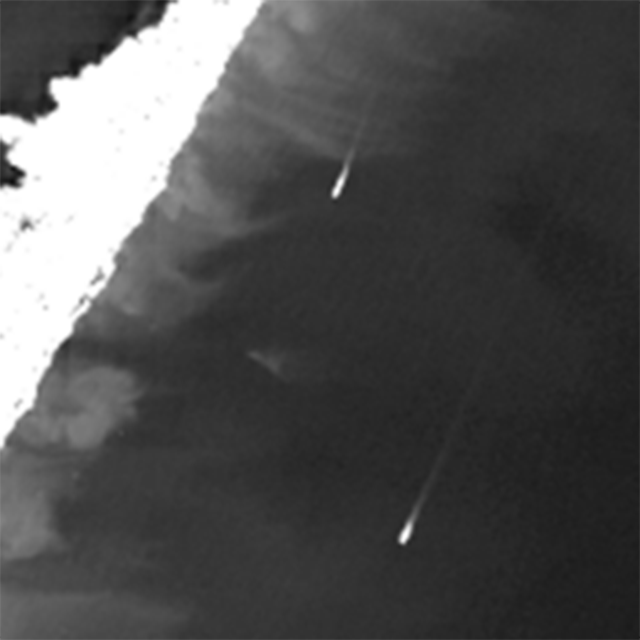}
 \end{minipage}
 \begin{minipage}[b]{.21\linewidth}
 \centering
 \includegraphics[scale=0.085]{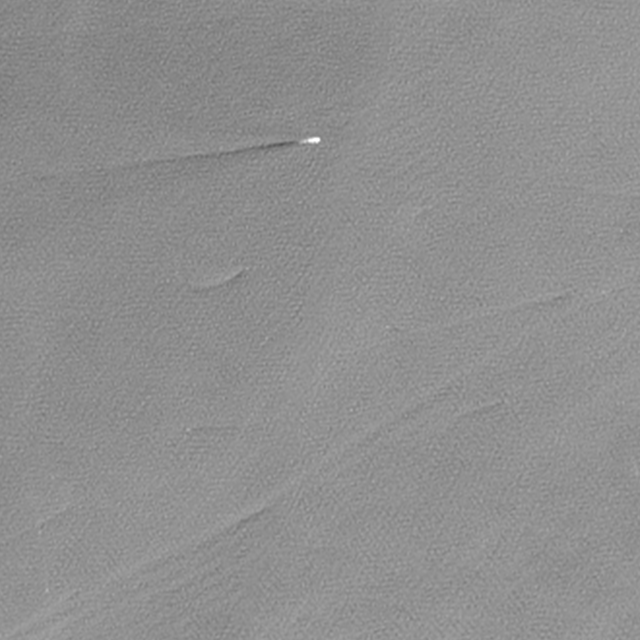}
 \end{minipage}
 \begin{minipage}[b]{.21\linewidth}
 \centering
 \includegraphics[scale=0.085]{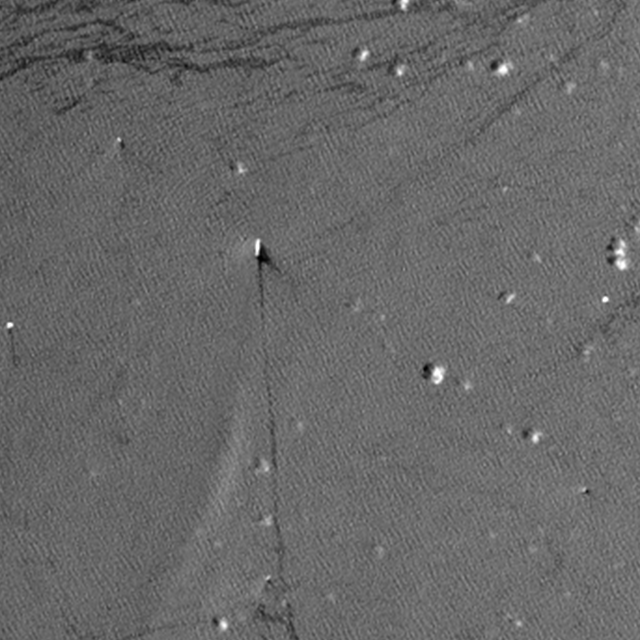}
 \end{minipage}
}
\subfigure[]
{
 \begin{minipage}[b]{.21\linewidth}
 \centering
 \includegraphics[scale=0.085]{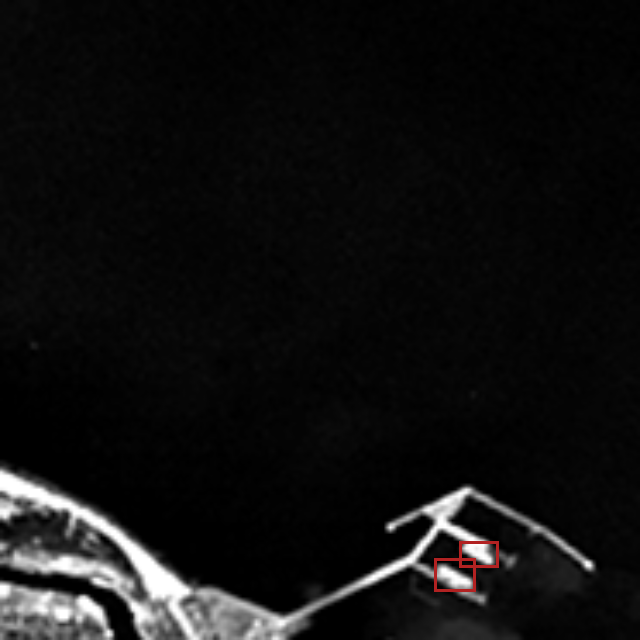}
 \end{minipage}
 \begin{minipage}[b]{.21\linewidth}
 \centering
 \includegraphics[scale=0.085]{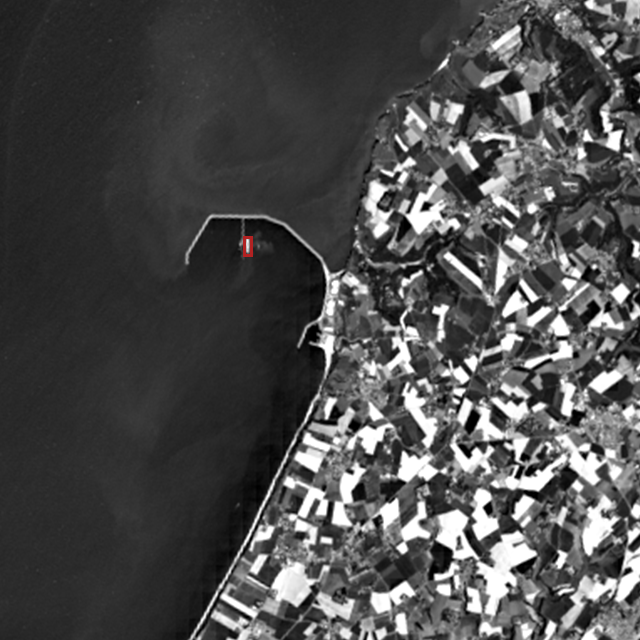}
 \end{minipage}
 \begin{minipage}[b]{.21\linewidth}
 \centering
 \includegraphics[scale=0.085]{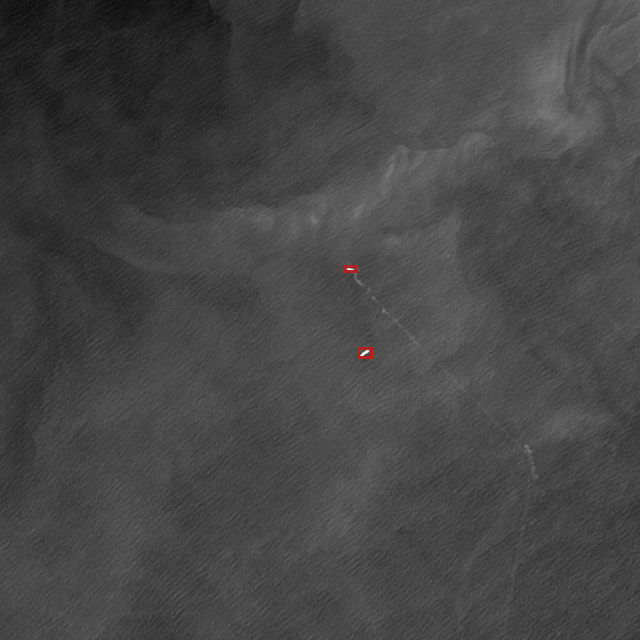}
 \end{minipage}
 \begin{minipage}[b]{.21\linewidth}
 \centering
 \includegraphics[scale=0.085]{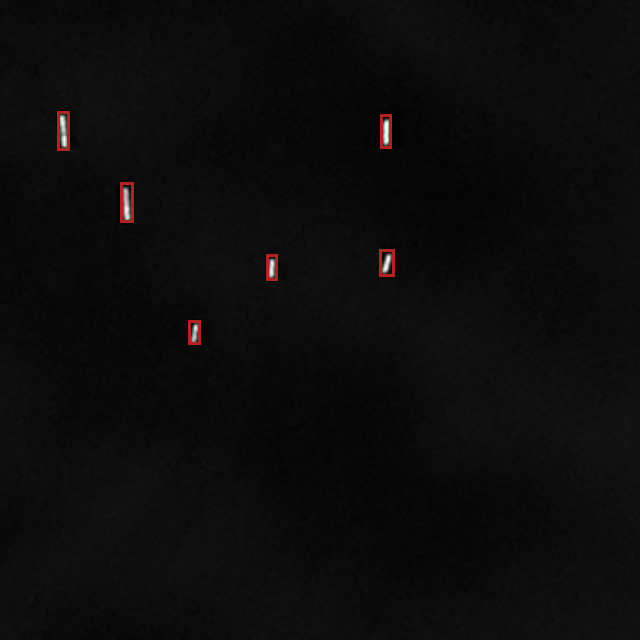}
 \end{minipage}
}
\subfigure[]
{
 \begin{minipage}[b]{.21\linewidth}
 \centering
 \includegraphics[scale=0.085]{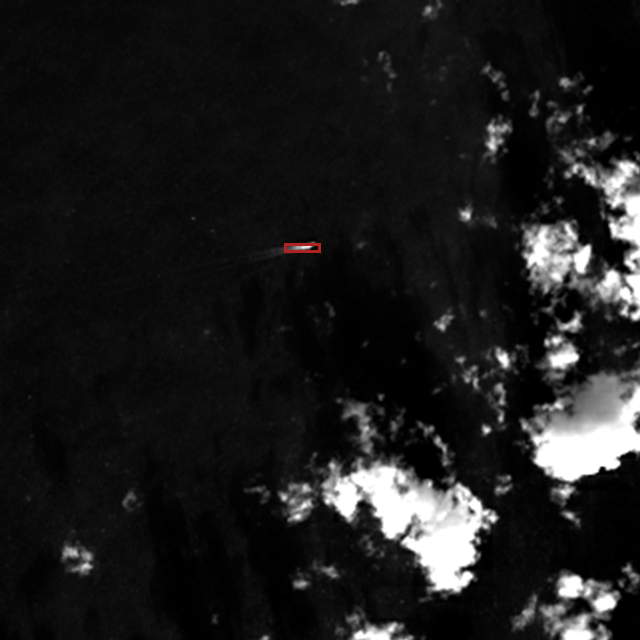}
 \end{minipage}
 \begin{minipage}[b]{.21\linewidth}
 \centering
 \includegraphics[scale=0.085]{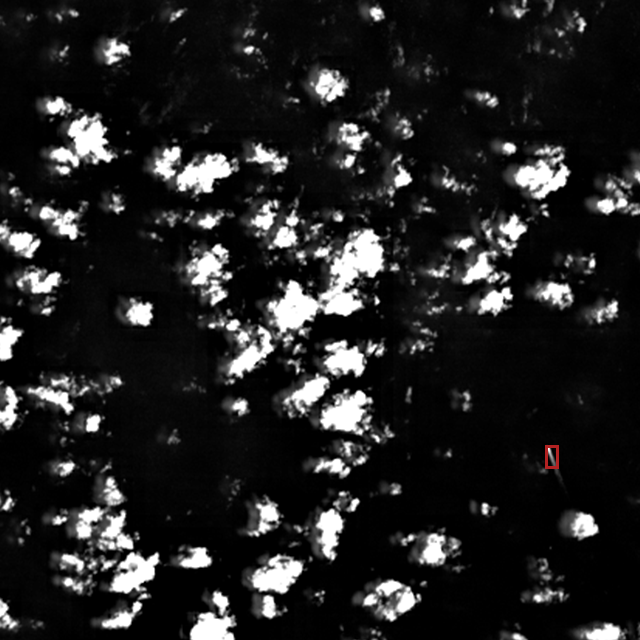}
 \end{minipage}
 \begin{minipage}[b]{.21\linewidth}
 \centering
 \includegraphics[scale=0.085]{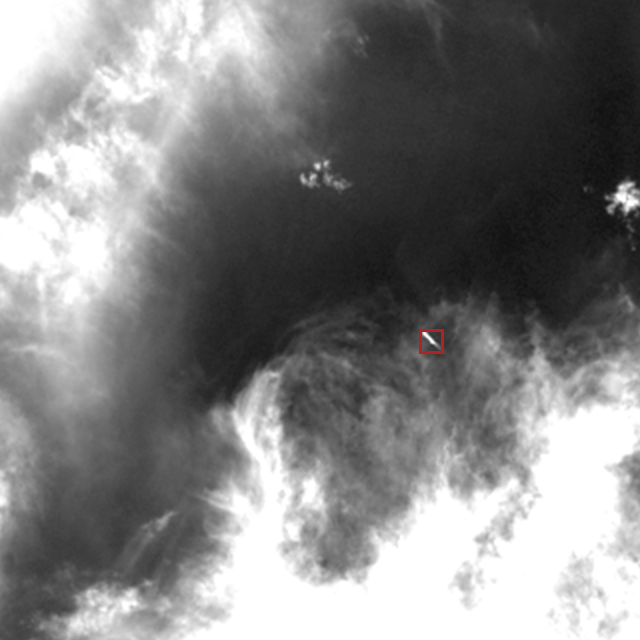}
 \end{minipage}
 \begin{minipage}[b]{.21\linewidth}
 \centering
 \includegraphics[scale=0.085]{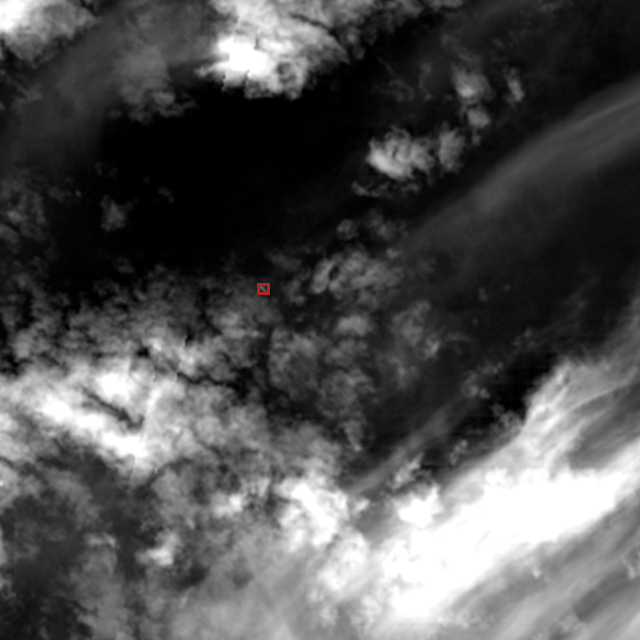}
 \end{minipage}
}
\subfigure[]
{
 \begin{minipage}[b]{.21\linewidth}
 \centering
 \includegraphics[scale=0.085]{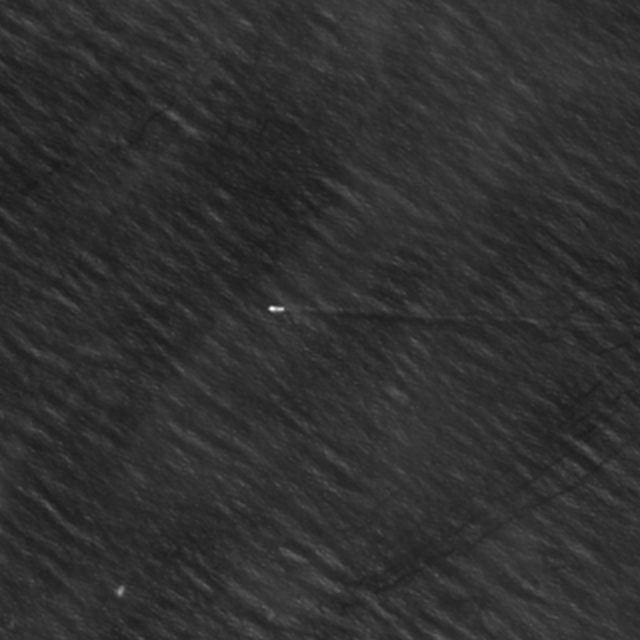}
 \end{minipage}
 \begin{minipage}[b]{.21\linewidth}
 \centering
 \includegraphics[scale=0.085]{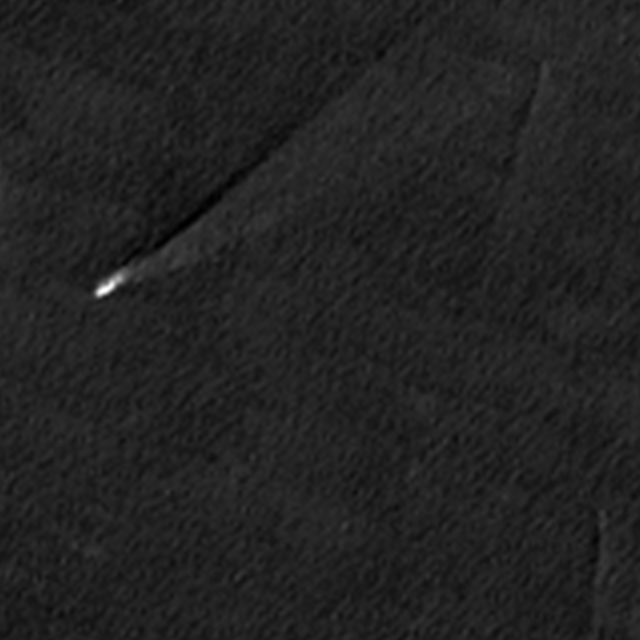}
 \end{minipage}
 \begin{minipage}[b]{.21\linewidth}
 \centering
 \includegraphics[scale=0.085]{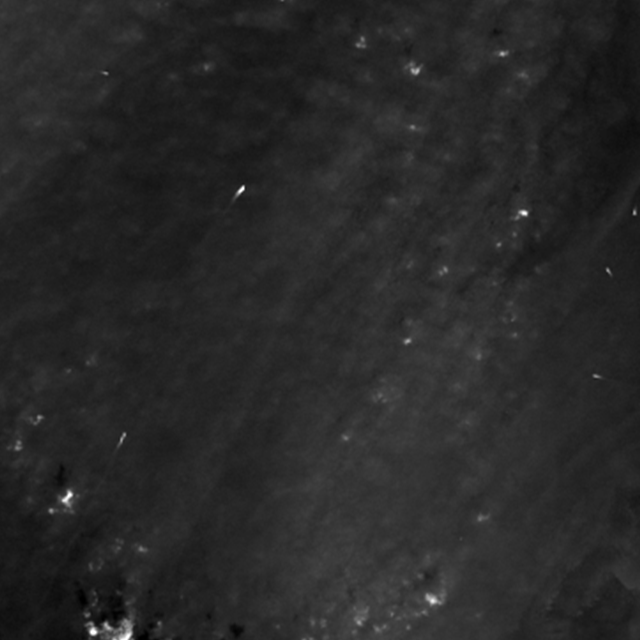}
 \end{minipage}
 \begin{minipage}[b]{.21\linewidth}
 \centering
 \includegraphics[scale=0.085]{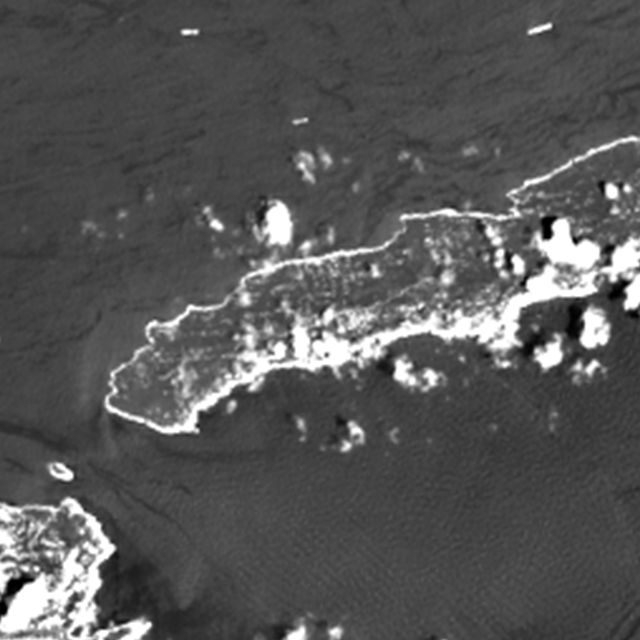}
 \end{minipage}
}
\caption{Diversity of scenery and weather: (a) ship with trails; (b) inshore and offshore scenes; (c) diverse clouds; and (d) sea wave.}
\label{comparsion_3}
\end{figure}
\begin{figure}
 \centering
 \includegraphics[scale=0.5]{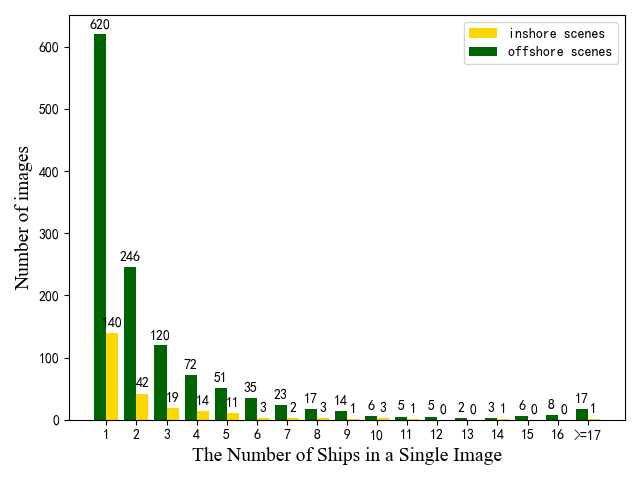}
 \caption{Statistics of nearshore and offshore ships are in each image.}
 \vspace{-0.5cm}
 \label{fig_6}
\end{figure}
\section {Infrared Ship Dataset with Scene Segmentation}
\subsection{Motation} 

In the network, we need the scene label to help the network learn the scene semantics. However, the public dataset offering the scene labels is rare. So, we propose a new dataset (IRSDSS), which contains not only the labels of targets but also the mask for lands and clouds.

\subsection{Dataset Construction}
Our dataset is based on images from Landsat8, generated after the Operational Land Imager (OLI) and Thermal Infrared Sensor (TIRS). The size of images in the datasets is $640 \times 640$. Our dataset includes 1491 images and 4062 targets from various backgrounds. The images are divided into training, validation, and testing sets in the 6:1:3 ratio.

\subsection{Properties of IRSDSS}

The proposed IRSDSS has the following features: 

\begin{itemize}
 \item [1)]
 \textit{Scene Segmentation:} Unlike conventional data annotations, IRSDSS offers ship labels and land and cloud masks, as shown in Fig. \ref{fig_12}.

 \item [2)]
 \textit{Target Size and Aspect Ratio:} The analysis of infrared ship targets in the dataset revealed lengths from 3.32 to 85.88 meters (mean: 15.53), areas from 16.28 to 4780.24 square meters (mean: 273.36), and aspect ratios from 0.25 to 4.20 (mean: 1.07), as detailed in Fig. \ref{comparsion_2}.

 \item[3)]
 \textit{Diverse Scenes:} IRSDSS consists of diverse and complex scenes. For instance, variations in the appearance of ship tails can be observed, influenced by factors such as the angle of incident sunlight and the roughness of the sea surface, as illustrated in Fig. \ref{comparsion_3}(a). In addition, the relationship between sea and land is diverse as Fig. \ref{comparsion_3}(b). The distribution of offshore and inshore ships is illustrated in Fig. \ref{fig_6}.

 \item[4)]
 \textit{Variable Weather:} Some weather conditions introduce harmful effects when detecting ships. Some clouds resemble the shape of ships, leading to false alarms, and thin clouds covering ships affect the contrast of ships, resulting in missings, as shown in Fig. \ref{comparsion_3}(c). Moreover, sea waves caused by winds, as shown in Fig. \ref{comparsion_3}(d), can also be recognized as ships, increasing the rate of false alarms.
\end{itemize}

\begin{figure*}
 \centering
 \includegraphics[scale=.16]{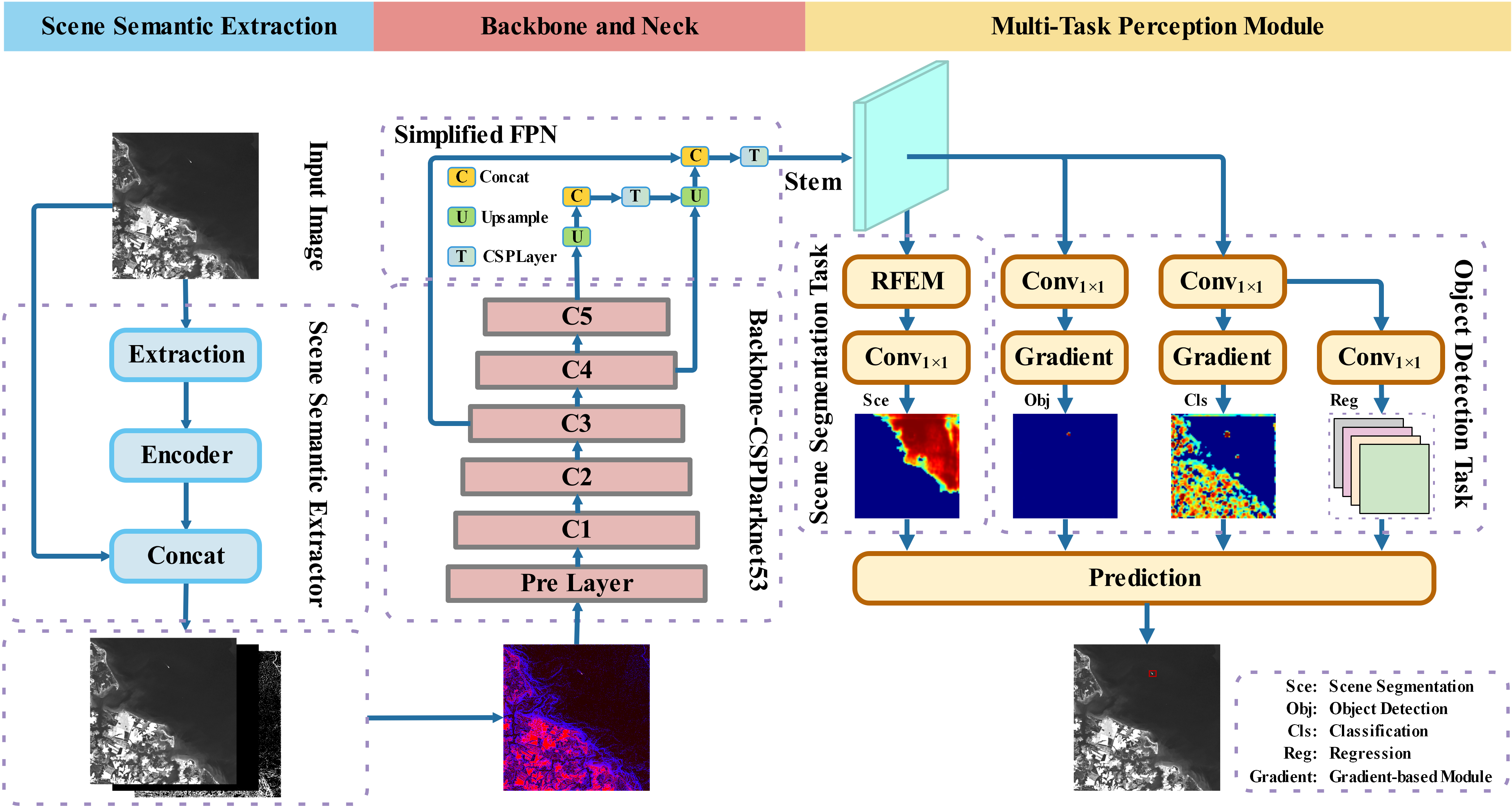}
 \caption{The SMPISD-MTPNet comprises three modules: 1) The SSE extracts scene semantics to enrich the inputs; 2) The CSPDarkNet-53 further processes these enhanced inputs; 3) The Multi-Task Perception Module analyzes the outputs of the CSPDarkNet-53 and predicts the result.}
 \label{backbone}
 \vspace{-0.5cm}
\end{figure*}

\section{Methodology}
SMPISD-MTPNet, which consists of three stages, is illustrated in Fig.\ref{backbone}. SSE enhances the input images. Then, CSPDarknet-53 \cite{1} extracts the deep features from the processed images, followed by simplified FPN fusing the deep features from various backbone layers. Finally, we introduce the Multi-Task Perception Module, which incorporates scene segmentation as an additional task, extending beyond the conventional tasks in prior networks. Furthermore, this module employs specially designed gradients to significantly enhance the semantics of dim and small targets.

\subsection{Scene Semantic Extraction}
Infrared ships lack high-level semantics. Traditional approaches \cite{6,30,31,35,36,47,48} could extract the semantic prior of each pixel based on expert knowledge. Due to the scenario-dependent nature of the ship, we could utilize the scene semantics to enrich the semantics of targets and guide the network. 
\begin{figure}
 \centering
 \includegraphics[scale=.3]{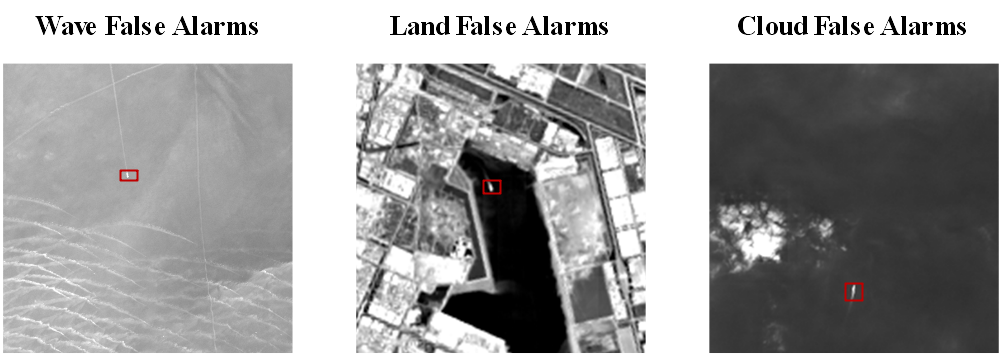}
 \caption{The scenes that cause false alarms.}
 \vspace{-0.5cm}
 \label{comparsion_4}
\end{figure}
\begin{figure}
 \centering
 \includegraphics[scale=.13]{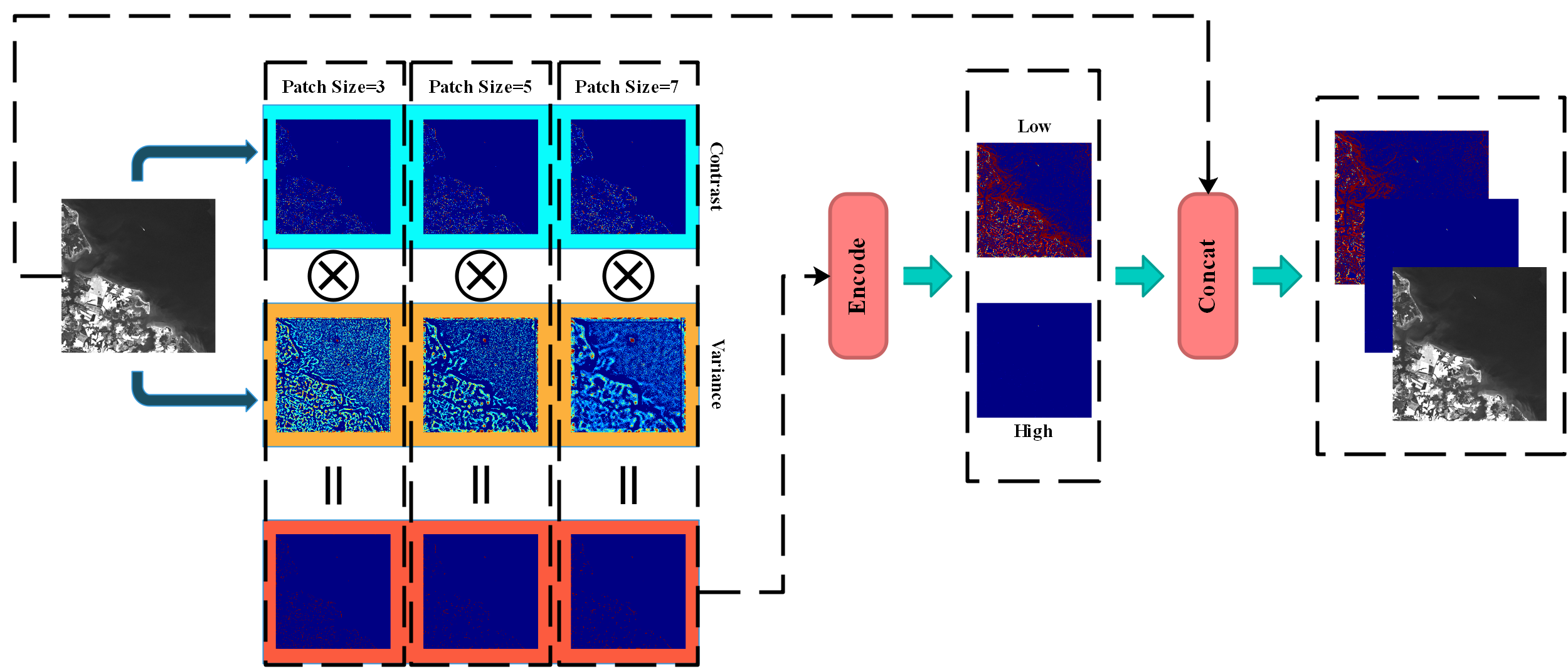}
 \caption{The process of SSE. Based on expert knowledge, the scene semantics in the original images are extracted, encoded, and concatenated with the original images.}
 \label{fig_8}
\end{figure}
\begin{figure}
\centering
\includegraphics[scale=.2]{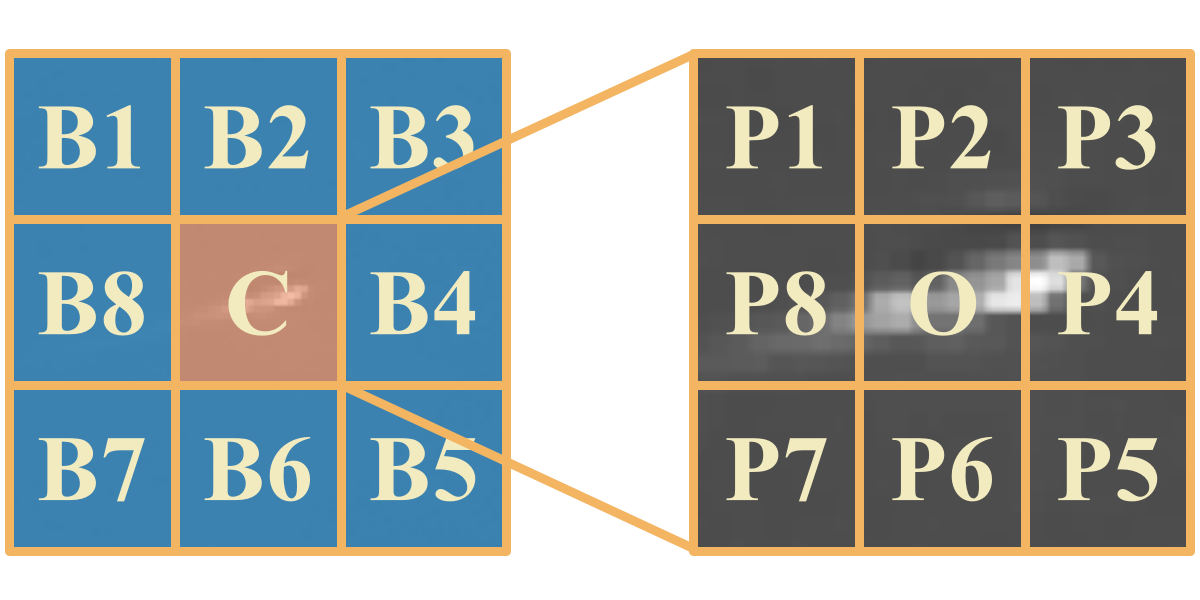}
\caption{The image patch is divided into a $3 \times 3$ grid with cells Bi (i = 1,...,8) and central cell C. Cell C is then subdivided into $3 \times 3$, including a central area O and peripheral areas Pi (i = 1,...,8).}
\label{fig3}
\end{figure}
As shown in Fig. \ref{comparsion_4}, the contrast of the infrared ship itself is higher than that of the background, and the variation of the background around ships is smaller than that of the land and clouds, where false alarms often appear. In addition, since the infrared ship size is variable over a wide range, we use multi-scale windows to extract features of targets with various sizes to increase the robustness of the module. According to the analysis above, we propose a module called SSE, as shown in Fig. \ref{fig_8}, and the process of SSE is as follows.

We use different windows to extract the features. We will illustrate the feature extraction process using a $j^{th}$ window as an example. We divide the window into nine blocks ($C, B_1, B_2, ..., B_8$), all of which are subdivided into the target patch ($O$) and background patch ($P_1, P_2, ..., P_8$), as shown in Fig. \ref{fig3}.

We then calculate each patch's average pixel value and pixel value variation. The formula is as follows:
\begin{align}
 \bar{I}^j_O &= \frac{ \sum_{(x,y)\in O} I_{x,y}}{(N^j_O)^2} \label{alg1_1}\\
 \bar{I}^j_{P_i} &= \frac{\sum_{(x,y)\in P_i} I_{x,y}}{(N^j_{P_i})^2} & (i = 1, \ldots, 8) \label{alg1_2}\\
 d^{j}_{O, P_i} &= \bar{I}^{j}_O - \bar{I}^{j}_{P_i} &(i = 1, \ldots, 8) \label{contrast_diff}\\
 V_R^{j} &=\frac{1}{8} \sum_{i=1}^{8} \left| d^{j}_{O, P_i} \right| & R\in(C, B1,\ldots, B8)\label{alg1_3}
\end{align}
where, $\bar{I}^j_O$ refers to the average pixel value of center patch $O$, which has size of $N_O^j \times N_O^j$, 
$\bar{I}^j_{P_i}$ denotes the average pixel value for each surrounding patches $P_i$ sized $N_{P_i}^j \times N_{P_i}^j$, 
$d^{j}_{O, P_i}$ represents the difference value between the surrounding patch $P_i$ and the central patch $O$, 
$V_R^{j}$ is defined as the mean absolute value of the differences between the central patch and the surrounding eight patches in block $R$, and $R$ can be the corresponding block for every pixel.
We compute the dissimilarity between the target and the background, defined by the following formulas:
\begin{align}
 D(O, P_i, P^{j}_{i+4}) &= d^{j}(O, P_i) \cdot d(O, P^{j}_{i+4}) \ (i = 1, \ldots, 4) \label{contrast_prod}\\
 S^j &= FS\left( 
 D^{j}(O, P_i, P_{i+4})
 \right) (i=1,...,4)
\end{align}
where, $D^j(O, P_i, P_{i+4})$ represents the product of the mean differences in two corresponding directions, and $S^j$ represents the dissimilarity value corresponding to the window the size of which is $N^j\times N^j$. To reduce the difference in the results for each pixel, we employ an algorithm, denoted as “$FS$”, designed to get the second largest value among values derived from four directions. We compute the $W^j$, which represents the variation of the background. 
\begin{align}
 W^j = \frac{V_C^{j}}{\sum_{i=1}^8 V_{B_i}^{j}}
 \label{w_alg}
\end{align}
Sequentially, we combine the variation of background and the local contrast. The final result $C^j$ is defined as follows:
 \begin{align}
 C^j = S^j \times W^j
 \label{C_alg}
\end{align}

We combine the results from different windows. Here, $C$ denotes the max value from different results $C^1, C^2, ..., C^L$ based on various windows.
 \begin{equation}
 C=\max\{C^1, C^2, \ldots, C^L\}
 \label{formula_0}
 \end{equation}

We are supposed to encode the results, and the process of encoding is defined below to enable the network to utilize the results more efficiently:
\begin{align}
 \label{eq_1_0} C'&=C\mod{\alpha_1}\\
 \label{eq_1_1} C''&=\frac{C}{\alpha_2}\\
 \label{eq_1_2} C_{out}&=\mathrm{Concat}(I, C', C'')
\end{align}
where, $\alpha_1$ and $\alpha_2$ serve as hyperparameters of encoding “mod” refers to applying the modulus operation to each pixel, and “Concat" means combining outputs and input. 

\begin{algorithm}
 \caption{Scene Semantic Extractor (SSE)}
 \KwData{Input Image $I$}
 \KwResult{Enhanced Dataset $C_{out}$ }
 \SetKwInOut{Input Image $I$}{Enhanced Dataset $C_{out}$}
 \For{$j = 1$ to $L$}
 {
 Divide the $j^{th}$ window into nine blocks: $C, B_1, \ldots, B_8$\;
 Each block is subdivided into a central patch $O$ and eight surrounding patches $P_1, \ldots, P_8$\;
 Compute the average pixel value for the central patch $O$ by (\ref{alg1_1})\;
 \For{$i = 1$ to $8$}
 {
 Compute the average pixel value for each surrounding patch $\bar{I}^j_{P_i}$ by (\ref{alg1_2})\;
 Calculate the mean absolute difference value for each patch $V_R^{j}\ (R \in {C, P_1,\hdots, P_8})$ by (\ref{contrast_diff}) and (\ref{alg1_3})\;
 }
 Calculate $D^j(O, P_i, P_{i+4})$ for corresponding patches by (\ref{contrast_prod})\;
 Get the second largest value $S^j$ from $\{D^j(O, P_i, P_{i+4}) | i = 1, 2, 3, 4\}$\;
 Compute the weighting factor $W^j$ by (\ref{w_alg})\;
 Calculate the result $C^j$ by (\ref{C_alg})\;
 }

Select the maximum value $C$ from ${C^1, C^2, \ldots, C^L}$\;
Map $C$ into $C'$ and $C''$ by (\ref{eq_1_0}) and (\ref{eq_1_1})\;
Concatenate $C'$ and $C''$ with $I$ by (\ref{eq_1_2})\ to get $C_{out}$;
\end{algorithm}

\subsection{Backbone and Neck}
\begin{figure}
 \centering
 \includegraphics[scale=.2]{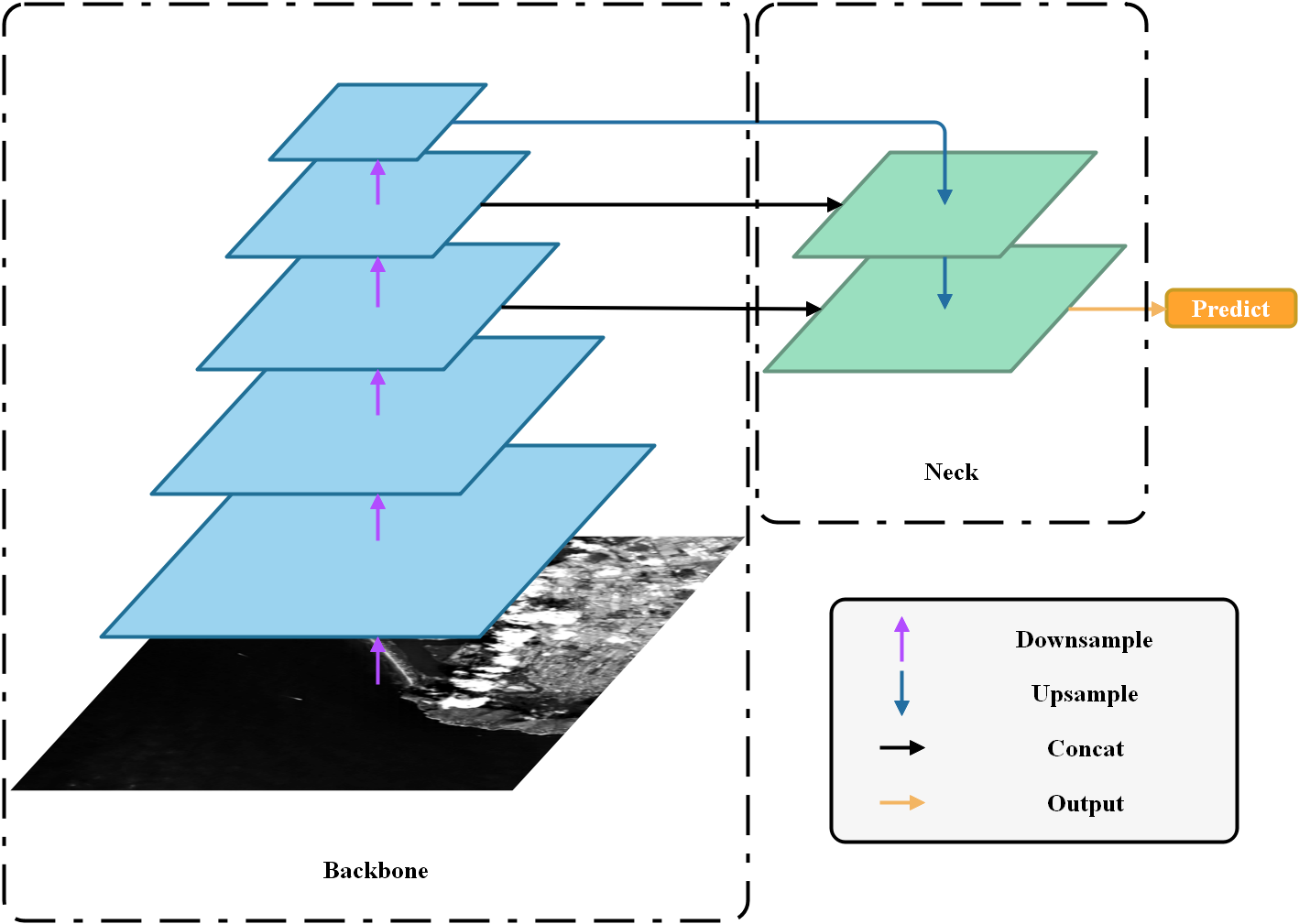}
 \caption{The backbone and the neck.}
 \label{fig_21}
\end{figure} 
We adopt CSPDarknet-53 as our backbone, extensively utilized in YOLOv4 \cite{49} and YOLOv5. In addition, we specifically utilize the $80 \times 80$ feature map for target detection. We employ a simplified FPN (shown in Fig. \ref{fig_21}) as the neck structure to enhance detection capabilities across targets of varying sizes.

\subsection{Multi-Task Perception Module}
\begin{figure}
 \centering
 \includegraphics[scale=.17]{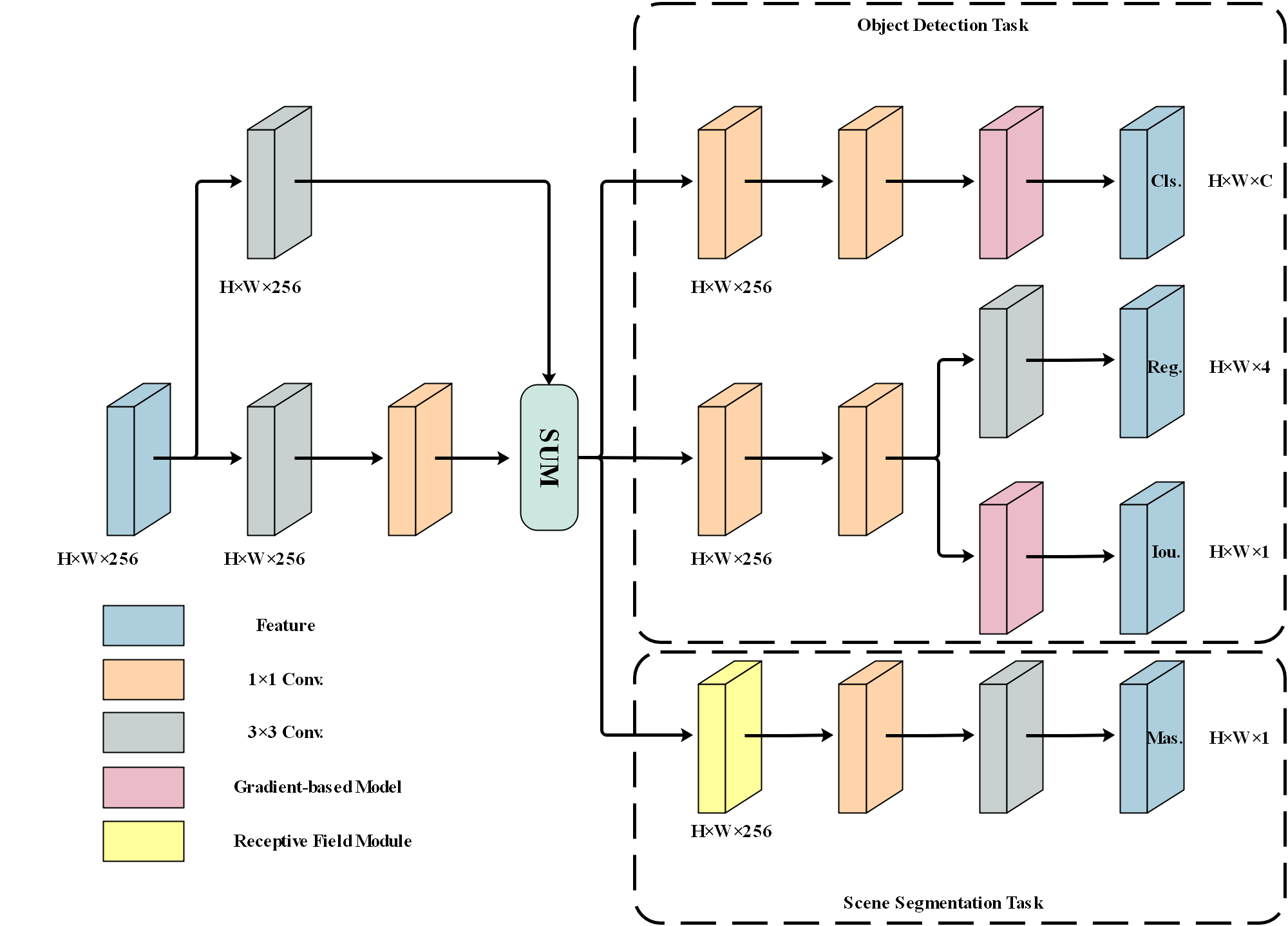}
 \caption{The structure of the Multi-Task Perception Module.}
 \label{fig_11_0}
\end{figure}

For infrared target detection, false alarms often arise from clouds and lands, and small targets often result in missings. We introduce scene segmentation to address the high false alarm rate. To enhance small target detection and suppress missings, we propose the Gradient-based Module. To prevent interference between the modules aimed at suppressing false alarms and improving recall rates, we introduce the Multi-Task Perception Module based on the decoupled head, which is widely used in detection network \cite{50,51,52}, as illustrated in Fig. \ref{fig_11_0}.

\subsubsection{Scene Segmentation Module}

\begin{figure}
 \centering
 \includegraphics[scale=.25]{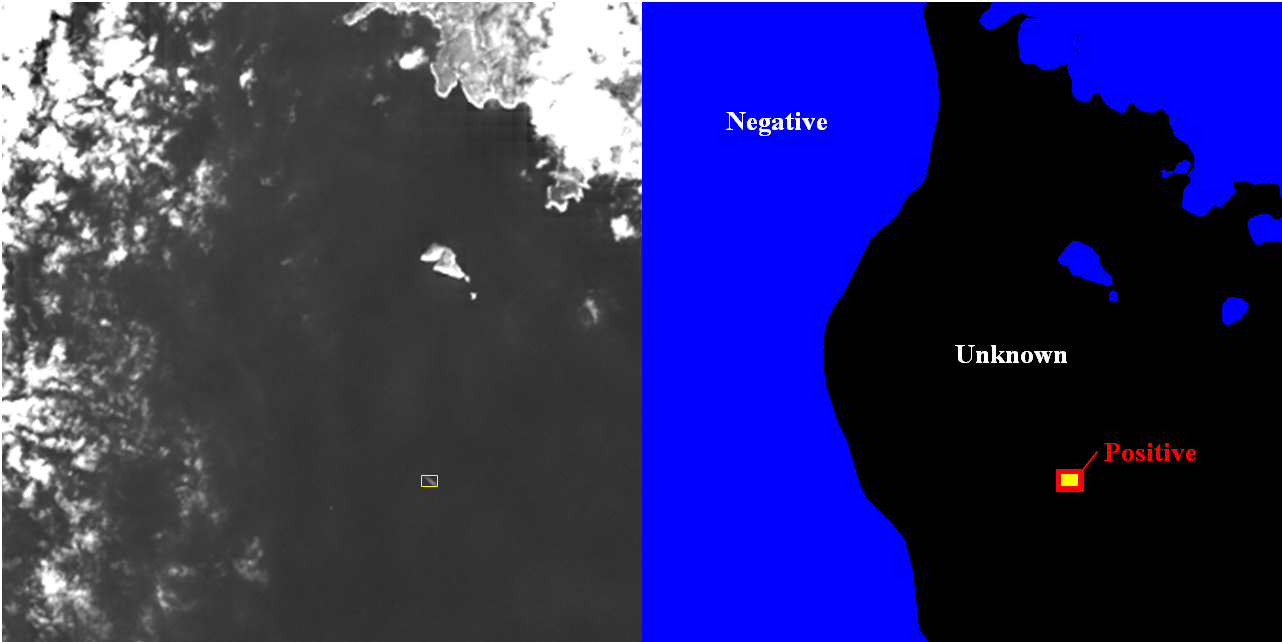}
 \caption{Blue represents the scene where the ships can't be, including land and clouds; the back denotes unknown areas; the red signifies positive samples generated by expanding the bounding boxes and twice their original size.}
 \label{fig_15}
\end{figure}

\begin{figure}
 \centering
 \includegraphics[scale=.08]{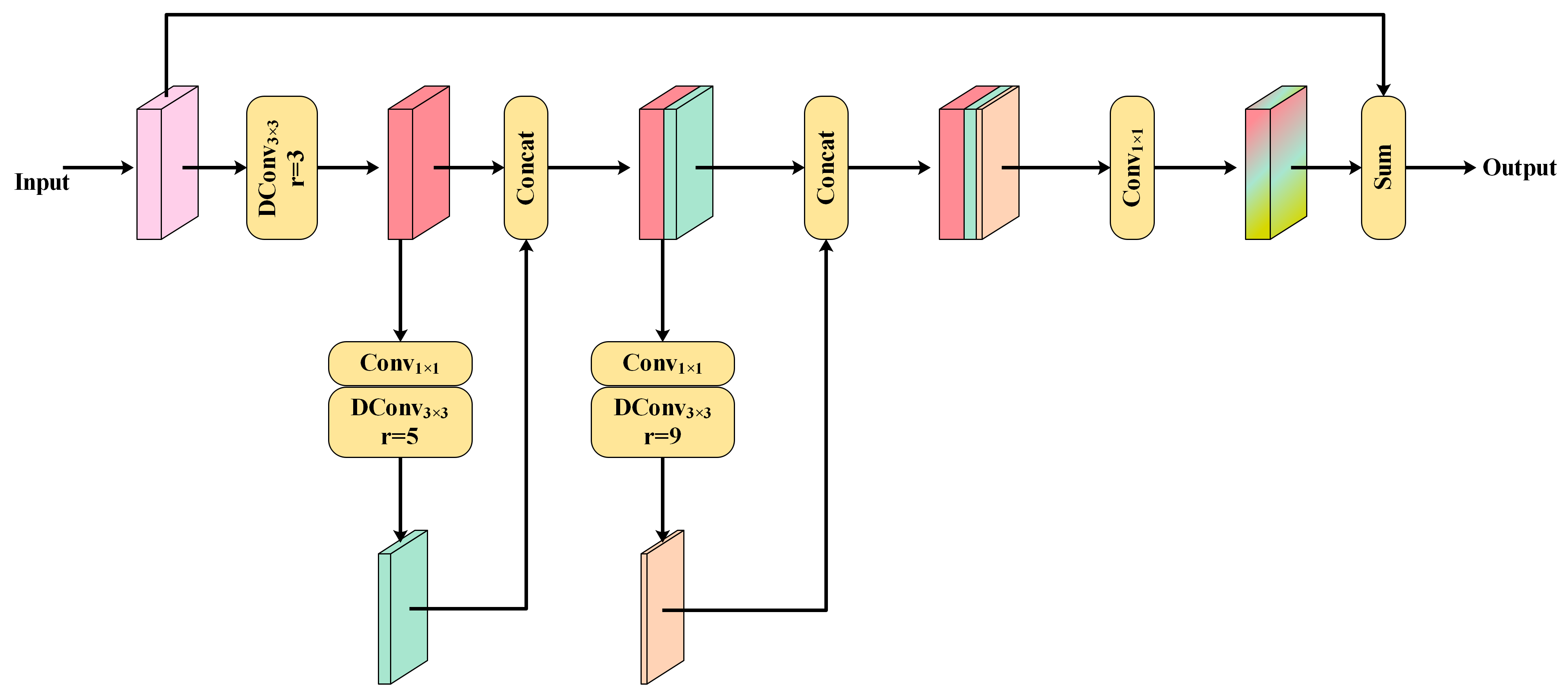}
 \caption{The structure of Receptive Field Expansion Module, where $DConv_{3\times 3}$ denotes the dilated convolution, r means the dilation rate of the correspondence dilated convolution, and $Conv_{1\times1}$ is the convolution with kernel whose size is 1.}
 \label{fig_20}
\end{figure}

Some false alarms are related to the scene. For example, the candidate targets detected on land must be false alarms. So, scene perception is vital for ship detection due to ships' high dependency. 
In addition to the conventional heads, which are integral for classifying, detecting, and regression, we add a new head responsible for scene semantic perception. When training the object detection head, we must assign positives and negatives to the prediction head. We define training samples using the Optimal Transport Assignment (OTA) \cite{53} algorithm. Training the scene semantic perception head also requires both positive and negative samples. Although IRSDSS annotates clouds and lands as separate categories for negative samples, we regard these scenes as a unified background category. For positive samples, drawing inspiration from SCRDet \cite{54}, and Knowledge-Driven Context Perception Network (KCPNet) \cite{13}, we generate masks from the ground-truth bounding boxes annotated in the dataset to serve as positive samples. Through experimental comparison, we have observed that enlarging the original ground-truth bounding boxes from the dataset by a factor of two before converting them into masks leads to significantly improved performance. Apart from the positive and negative samples previously discussed, we classify the other areas as unknown and exclude them from the environmental awareness detection head training. Creating unknown regions in training can provide redundancy in prediction, enhance generalization and reduce the negative impact of data labelling errors. The conversion of annotated detection boxes to masks is depicted in Fig. \ref{fig_15}. Adequate scene perception depends on a broad receptive field. Therefore, expanding the receptive field is essential to improving visual analysis capabilities. So, we introduce the Receptive Field Expansion Module, as shown in Fig. \ref{fig_20}.

\subsubsection{Gradient-based Module}
\begin{figure*}
 \centering
 \includegraphics[scale=.16]{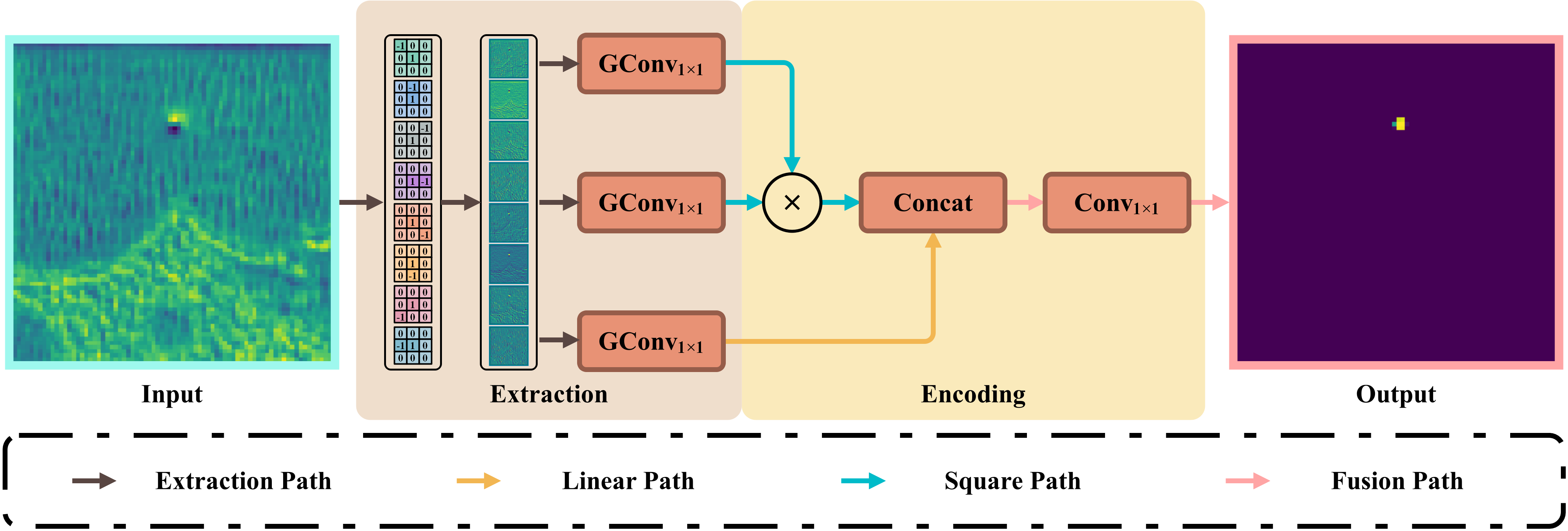}
 \caption{The figure shows the entire process of the Gradient-based Module. $GConv _{1 \times 1}$ refers to group convolution with a kernel size of 1 and group value of 8. $Conv_{1\times 1}$ denotes $1\times 1$ convolution used for channel fusion.}
 \label{fig_19}
\end{figure*}

Some infrared ships are small and dim, and we can not depend on their features to detect them, leading to missings. So, we use the difference between targets and backgrounds to detect the target. Inspired by the Gradient-Guided Learning Network (GGL-Net) \cite{55}, we design a module which is based on gradients to detect targets, as illustrated in Fig. \ref{fig_19}.  

\begin{figure}
 \centering
 \includegraphics[scale=.13]{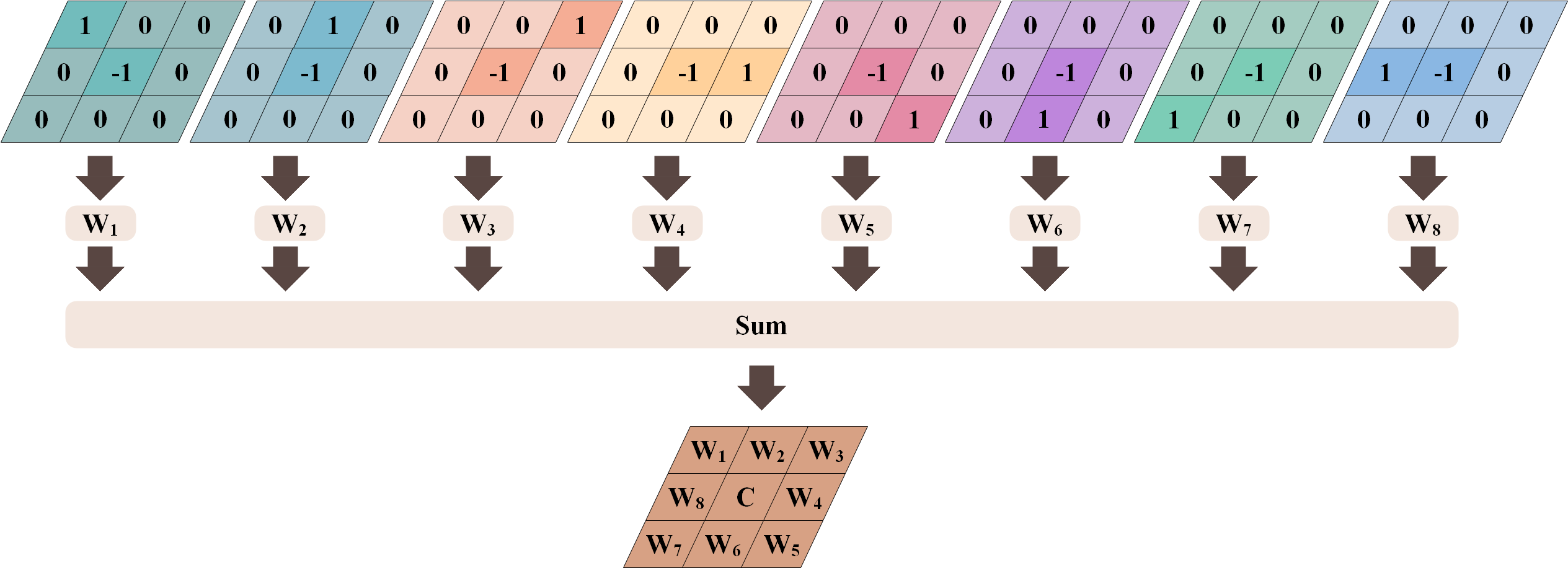}
 \caption{Almost gradient operators can allocate different weights to the gradient operators in eight directions and then sum these gradient operators. $W_i$ represents eight different weights for eight directions, and $C=-\sum_{i=1}^8 W_i$.}
 \label{fig_18}
\end{figure}

This module consists of two stages: extraction and encoding. We use gradient operators to get the difference between targets and backgrounds. Unlike conventional modules that rely on predefined gradient operators in the gradient extraction phase, we adopt a neural network to learn the weights of various gradient operators. This approach allows for a more comprehensive and flexible representation of target changes by weighting combinations of gradient operators from eight directions, as shown in Fig. \ref{fig_18}. 

Then, we extract the difference between the local features and their surroundings and use group convolution to weigh and fuse the results from eight directions. 

We encode the values and output the result. Inspired by Attentional Local Contrast Networks (ALCNet) \cite{14} and MLCL \cite{16}, we encode the gradients not only by linear encoding but also by square encoding.

In conclusion, we have analyzed the causes of false positives and negatives. We incorporate the scene segmentation module as an additional task to reduce false positives and utilize the Gradient-based Module to decrease the occurrence of false negatives, particularly for objects of small size.

\subsection{Soft Fine-tuning}
In order to improve performance, this research employs data augmentation techniques, including Mosaic \cite{49}, MixUp \cite{56}, and affine transformations on data pre-processed by the SSE.

Typically, strong data augmentation can introduce distortions. So, we have to stop augmentation in the final training phases, allowing us to fine-tune the network with accurate data for better adaptability, which would suppress the improvements of data augmentation. So, we propose a new data augmentation, Soft Fine-tuning, where the following formula defines the proportion of augmented images across various epochs.
\begin{equation}
 R=1-\beta \times(\frac{m}{M})^2 (\beta\in(0,1))
\end{equation}
where, R denotes the ratio of augmented images, m denotes the epoch number within the training sequence, and M denotes the total number of epochs. $\alpha$ adjusts the rate at which augmented image proportion decreases. This training strategy preserves the generalization ability induced by data augmentation while refining the network with actual data inputs.

\subsection{Loss Function}
According to the previous component, the conventional object prediction head, our network uses image segmentation for scene perception. When introducing the new tasks, the loss function is shown as follows:

\begin{equation}
\begin{aligned}
L=&\frac{\lambda _1}{N_f}\sum_{n=0}^{N_f}{L_{iou}\left( \widehat{x}_n,\widehat{y}_n,\widehat{w}_n,\widehat{h}_n,x_n,y_n,w_n,h_n \right)}
\\
&+\frac{\lambda _2}{N_f}\sum_{n=0}^{N_f}{L_{obj}}\left( \widehat{u}_n,u_n \right) +\frac{\lambda _3}{N_f}\sum_{n=0}^{N_f}{L_{cls}}\left(\widehat{t}_n ,p_n \right)
\\
&+\frac{\lambda _4}{N_m}\sum_{n=0}^{N_m}{L_{mas}}\left( \widehat{m}_n,m_n \right) 
\end{aligned}
\end{equation}
where, $N$ denotes the total number of prediction boxes. $N_f$ and $N_m$ are the number of positive and negative samples for object and false alarms. The predicted bounding boxes are denoted by $ (\widehat{x}_n, \widehat{y}_n, \widehat{w}_n, \widehat{h}_n ) $, while the ground truth bounding boxes are represented by $(x_n, y_n, w_n, h_n)$. The confidence score of each predicted bounding box, represented by $u_n$, indicates the probability of the target. The variable $\widehat{u_n}$ signifies the confidence score associated with each predicted bounding box, and $u_n$ is the existence of a true target. The variable $p_n$ is the intersection over the Union (IoU) value between positive samples within ground truth bounding boxes and matching predicted counterparts. $t_n$ indicates the confidence level of these predicted bounding boxes regarding the classification of their respective ground truth categories. The predicted mask is indicated by $\widehat{m}_n$, while $m_n$ represents the actual mask. $\lambda_1$, $\lambda_2$, $\lambda_3$, and $\lambda_4$ are hyperparameters. $L_{cls}$, $L_{mas}$, and $L_{obj}$ are cross-entropy loss, and $L_{iou}$ represents the IoU loss function.

\section{Experiment}
The section opens with a detailed presentation of the experimental setup, encompassing the environment, hyperparameter settings, and evaluation metrics. Following this, we conduct ablation studies on the network components proposed in this research to evaluate their functionality and impact. Ultimately, we conduct a performance comparison between the proposed SMPISD-MTPNet and other networks on IRSDSS. 
\subsection{Experiment Settings}

This section provides an exhaustive overview of the experiment details, covering the experiment environment, settings, and the metrics used to evaluate network performance.
\begin{table*}[h!]
 \centering
		\setlength{\belowcaptionskip}{4mm} 
		\captionsetup{font={footnotesize,stretch=1.25},justification=raggedleft}
		\caption{\fontsize{10.5bp}{17bp}{Ablation Studies of Each Component In SMPISD-MTPNet}}
 {\fontsize{14pt}{14pt}\selectfont 
 \scalebox{0.65}{ 
 \renewcommand{\arraystretch}{1.05}
 \begin{tabular}{c|c|c|c|c|c|c|c|c|c|c}
 \Xhline{2pt}
 \makebox[0.15\textwidth]{\multirow{2}{*}{Net Configure}} 
 &\multicolumn{4}{c|}{Net component} 
 & \makebox[0.12\textwidth]{\multirow{2}{*} {$AP_{all}^{@50:95}$}}
 &\makebox[0.10\textwidth] {\multirow{2}{*} {$AP_{all}^{@50}$}} 
 & \makebox[0.10\textwidth]{\multirow{2}{*} {$AP_{all}^{@75}$}}
 & \makebox[0.10\textwidth]{\multirow{2}{*} {$AP_S^{@50}$}} 
 & \makebox[0.10\textwidth]{\multirow{2}{*} {$AP_M^{@50}$}} 
 & \makebox[0.10\textwidth]{\multirow{2}{*} {$AP_L^{@50}$}} \\

 \Xcline{2-5}{0.4pt}
 & \makebox[0.1\textwidth]{Pre} 
 & \makebox[0.1\textwidth]{Soft} 
 & \makebox[0.1\textwidth]{Gradient} 
 &\makebox[0.1\textwidth]{Scene} & & & & & & \\
 \Xhline{1pt}
 
Baseline & - & - & - & - & 42.3 & 90.9 & 30.3 & 90.2 & 96.8 & 88.5 \\
Net1 & \checkmark & - & - & - & 40.1 & 91.9 & 27.2 & 90.5 & 96.1 & \textbf{93.1} \\
Net2 & - & \checkmark & - & - & 41.8 & 92.1 & 29.6 & 92.0 & 96.4 & 88.4 \\
Net3 & - & - & \checkmark & - & 40.9 & 91.6 & 29.2 & 90.6 & 95.5 & 89.9 \\
Net4 & - & - & - & \checkmark & 40.6 & 91.2 & 27.5 & 89.6 & 95.4 & 90.0 \\
Net5 & \checkmark & \checkmark & - & - & 42.5 & 92.8 & 32.8 & 91.9 & 96.3 & 89.7 \\
Net6 & \checkmark & - & \checkmark & - & 42.4 & 91.9 & 30.9 & 91.3 & 95.7 & 90.8 \\
Net7 & \checkmark & - & - & \checkmark & 42.6 & 92.5 & 31.3 & 92.1 & 96.2 & 87.1 \\
Net8 & -& \checkmark & \checkmark& - & 43.3 & 93.0 & 34.2 & \uline{93.1} & 96.6 & 90.1 \\
Net9 & -& \checkmark & - & \checkmark & 43.0 & 91.5 & 33.6 & 90.6 & 95.6 & 91.5 \\
Net10 & - & - & \checkmark & \checkmark & 40.9 & 91.2 & 29.6 & 91.3 & 94.8 & 87.3 \\
Net11 & \checkmark & \checkmark & \checkmark & - & 43.3 & \uline{93.8} & 32.3 & 93.0 & 97.1 & 92.5 \\
Net12 & \checkmark & \checkmark & - & \checkmark & \textbf{44.6} & 93.4 & \textbf{37.1} & 92.1 & \textbf{97.6} & \uline{93.0} \\
Net13 & \checkmark & - & \checkmark & \checkmark & 41.3 & 92.4 & 30.4 & 91.3 & 97.1 & 90.7 \\
Net14 & - & \checkmark & \checkmark & \checkmark & 44.2 & 92.7 & \uline{36.7} & 92.8 & 96.8 & 86.5 \\
Net15 & \checkmark & \checkmark & \checkmark & \checkmark & \uline{44.5} & \textbf{94.5} & 34.6 & \textbf{93.6} & \uline{97.3} & 92.2 \\
\Xhline{2pt}
 \end{tabular}
 }}
\label{comparsion_tablet}
\vspace{2pt}
\noindent
\setstretch{.8}
{\small \\
Baseline refers to the network without any modules proposed in this paper, and Neti (for i = 1 to 15) denotes the network employing various modules proposed in this research. Pre, Soft, Gradient, and Scene mean SSE, Soft Fine-tuning, Gradient-based Module, and Scene Segmentation Module.}
\end{table*}
\begin{itemize}
 \item [1)]
 \textit{Experimental Environment:}
 All experiments described herein were conducted on a uniform server configuration. This server runs on Ubuntu 20.04 LTS and is powered by an Intel i5-13490F CPU and an Nvidia RTX 4080 GPU with 16 GB of memory. Furthermore, the SMPISD-MTPNet is based on the PyTorch framework.
 \item[2)]
 \textit{Experiment Settings:}
 This research employs the experimental foundation of the IRSDSS dataset, containing 1491 images. Training is conducted with images in float32 format, divided into 64 batches of 16 images each. The study is designed to run for 150 epochs, totalling 9600 iterations. Initially, the learning rate progressively rises from 0 to 0.2 over the first 64 iterations, after which it is modulated via CosineAnnealingLR. The exponential moving average (EMA) approach is adopted to ensure network robustness further.
 \item[3)]
 \textit{Data Augmentation:}
 In this research, we adopted Mosaic, Mixup, and affine transformation as data augmentation techniques with randomly generated hyperparameters to simulate changes. Precisely, Mosaic and Mixup technologies dynamically adjust the stretching of the images range to 0.5 to 1.5 times the original size. At the same time, the affine transformation includes translation in the range of -64 to 64 pixels, rotation from -10 to 10 degrees and -2 to 2 degrees of miscutting on the x- and y-axes. Integrating these data augmentations enriches the training on data diversity, enhancing the adaptability of the module to new data.
 \item[4)]
 \textit{Evaluation Metrics:} 
 We select widely accepted COCO metrics, $AP_{all}^{@50}$, $AP_{all}^{@50:95}$, $AP_S^{@50}$, $AP_M^{@50}$, $AP_M^{@50}$, and $AP_{all}^{@75}$, to serve as our evaluation metrics. The IoU threshold set for average precision (AP) calculation in the research is 0.5.
\end{itemize}
\subsection{Ablation Studies}
\begin{figure*}
\includegraphics[scale=.3]{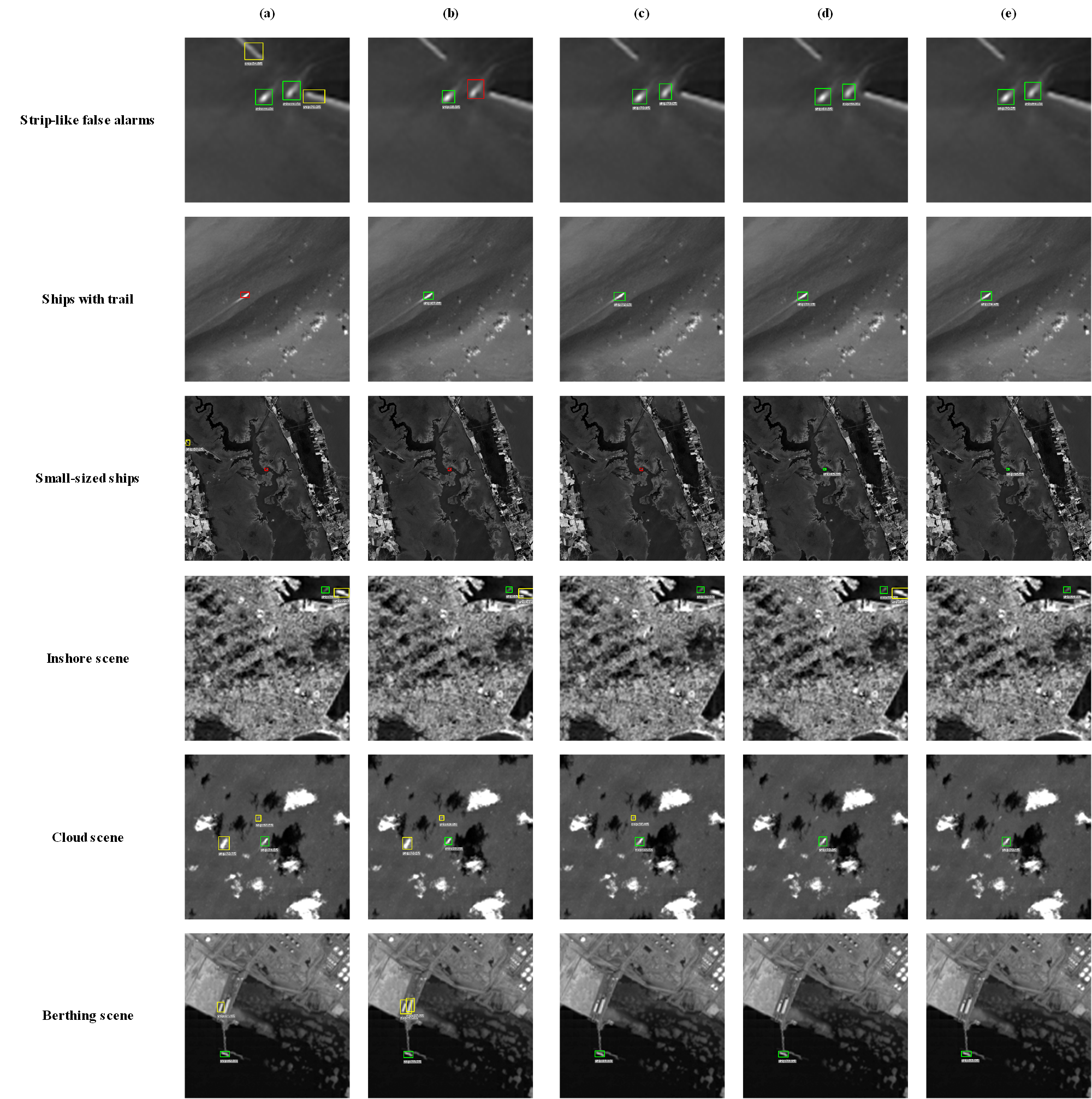}
\caption{This figure illustrates the effectiveness of three distinct modules and a training strategy proposed in this research through ablation experiments across six scenarios within IRSDSS. The visualization uses green, yellow, and red outlines to categorize ships correctly identified, overlooked ships, and incorrect detections, respectively. These networks include (a) the baseline, (b) the initial network with SSE(Net 1),(c) the initial network with SSE and Soft Fine-tuning, (d) the baseline network with SSE, Soft Fine-tuning and Gradient-based Module, (e) the base network with SSE, Soft Fine-tuning, Gradient-based Module, and Scene Segmentation Module.}
\label{fig_16}
\end{figure*}

In this section, we use many ablation experiments to explore the effects of the components and training strategy proposed in this research. Initially, we assessed the performance of the primary network without employing each component described in this research. Subsequently, we experimented with the impact of each module or training strategy on individual network performance. Following this, we combined pairs or trios of modules to observe the interplay between each module. Finally, we evaluated the performance of our network with the application of all modules described in this research. The results obtained from these experiments are listed in Table \ref{comparsion_tablet}. Fig. \ref{fig_16}. compares example results in different scenarios.
	\begin{table}[h!]
		\centering
		\setlength{\belowcaptionskip}{4mm} 
		\captionsetup{font={footnotesize,stretch=1.25},justification=raggedleft}
		\caption{\fontsize{10.5bp}{17bp}{Comparison of Scene Semantic Extraction}}
		\fontsize{14pt}{14pt}\selectfont
		\scalebox{0.65}{ 
			\renewcommand{\arraystretch}{1.05}
			\begin{tabular}{c|c|c|c|c|c|c|c}
				\Xhline{2pt}
				\makebox[0.05\textwidth]{\multirow{2}{*}{Name}} 
				&\makebox[0.06\textwidth]{\multirow{2}{*}{Features}}
				&\makebox[0.08\textwidth]{\multirow{2}{*} {$AP_{all}^{@50:95}$}}
				&\makebox[0.055\textwidth] {\multirow{2}{*} {$AP_{all}^{@50}$}} 
				&\makebox[0.055\textwidth]{\multirow{2}{*} {$AP_{all}^{@75}$}}
				& \makebox[0.055\textwidth]{\multirow{2}{*}  {$AP_S^{@50}$}}  
				& \makebox[0.055\textwidth]{\multirow{2}{*} {$AP_M^{@50}$}}  
				& \makebox[0.055\textwidth]{\multirow{2}{*} {$AP_L^{@50}$}} \\
				
			 & & & & & & & \\
				\Xhline{1pt}
				
				F1 & Canny & 43.9 & 93.4 & 35.6 & 93.0 & \textbf{97.4} & 89.4 \\
				F2 & MPCM & 45.0 & 93.7 & \textbf{36.8} & 92.4 & 97.0 & \textbf{94.0} \\
				F3 & $SSE^1$ & \textbf{44.6} & \uline{93.8} & 35.2 & \uline{93.2} & 97.2 & \uline{92.2} \\
				F4 & $SSE^2$ & 43.8 & 92.9 & \uline{36.1} & 91.8 & 96.7 & 91.2 \\
				F5 & $SSE^3$ & \uline{44.5} & \textbf{94.5} & 34.6 & \textbf{93.6} & \uline{97.3} & \uline{92.2} \\
		\Xhline{2pt}
		\end{tabular}}
\label{FPM}
\vspace{2pt}
\noindent
\setstretch{.8}
{\small
\\
Fi (i=1 to 5) represents experimental groups using multiple feature extraction methodologies. SSE$^1$ specifically employs the low bit of the SSE outcome, SSE$^2$ exclusively uses the high bit, and SSE$^3$ incorporates both the high and low bits of the SSE result.
}
\end{table}

\subsubsection{Effect of Scene Semantic Extraction}
As shown in Table \ref{comparsion_tablet}, after the introduction of scene semantic extraction (Net1), $AP_{all}^{@50}$ increased from 90.9\% of the baseline network to 91.9\% of the network. Similarly, the comparison between the Net14 and the final network (Net15) shows that almost all evaluation metrics values have improved after introducing scene semantic extraction. $AP_L^{@50}$ increased from the original 86.5\% to 92.2\%, an increase of 5.7\%. 
Based on Table \ref{FPM}, to evaluate the effectiveness of the SSE proposed in network performance research, we conducted comparative experiments against traditional feature extraction modules, as evidenced by the analysis of groups F1, F2, and F5. 
Analysis of experiment groups F1 and F5 from Table \ref{FPM} reveals that the utilization of edge features shows an unsatisfactory performance for large targets.  
Comparing group F2 with group F5, which utilizes MPCM to extract local contrast, is less effective than SSE. 
As mentioned earlier, SSE necessitates encoding the result to capture both high and low bits of the value. 
Comparison among F3, F4, and F5 reveals that independently using either the high or low bits of the values does not effectively enhance the network's performance. 

\begin{table}[h!]
 \centering
 \setlength{\belowcaptionskip}{4mm} 
 \captionsetup{font={footnotesize,stretch=1.25},justification=raggedleft}
 \caption{\fontsize{10.5bp}{17bp}{Comparison of Gradient-guided Module}}
 \fontsize{14pt}{14pt}\selectfont 
 \scalebox{0.65}{ 
 \renewcommand{\arraystretch}{1.05}
 \begin{tabular}{c|c|c|c|c|c|c|c}
 \Xhline{2pt}
 \makebox[0.05\textwidth]{\multirow{2}{*}{Name}} 
 &\makebox[0.06\textwidth]{\multirow{2}{*} {Module}}
 &\makebox[0.08\textwidth]{\multirow{2}{*} {$AP_{all}^{@50:95}$}}
 &\makebox[0.055\textwidth] {\multirow{2}{*}{$AP_{all}^{@50}$}} 
 &\makebox[0.055\textwidth]{\multirow{2}{*} {$AP_{all}^{@75}$}}
 & \makebox[0.055\textwidth]{\multirow{2}{*} {$AP_S^{@50}$}} 
 & \makebox[0.055\textwidth]{\multirow{2}{*} {$AP_M^{@50}$}} 
 & \makebox[0.055\textwidth]{\multirow{2}{*} {$AP_L^{@50}$}}  \\
 
 & & & & & & & \\
 \Xhline{1pt}
 
 D1 & FENetwVW& 44.2 & 93.2 & 35.5& 92.8 & 96.0 & 91.7 \\
 
 D2 & MLCL & 44.1 & 93.0 & \uline{36.3} & 92.2 & 97.0 & \textbf{92.3} \\
 D3 & $GM^1$ & \textbf{44.9} & \uline{93.4} & \textbf{36.9} & \uline{93.2} & 96.7 & 90.2 \\
 D4 & $GM^2$ & 44.2 &93.3 &35.2 & 92.1 & \textbf{97.5} & 90.5 \\
 D5 & $GM^3$ & \uline{44.5} & \textbf{94.5} & 34.6 & \textbf{93.6} & \uline{97.3} & \uline{92.2} \\
 \Xhline{2pt}
 \end{tabular}}
\label{Dual}
\vspace{2pt}
\noindent
\setstretch{.8}
{\small \\
Di (i=1 to 5) denotes the research groups that utilize gradient processing-based modules. GM$^1$ indicates the exclusive use of linear gradients in the Gradient-based Model, GM$^2$ represents the use of square gradients, and GM$^3$ involves the combined use of both linear and square gradients in the Gradient-based Module. }
\end{table}
 
\begin{table}[h!]
		\centering
		\setlength{\belowcaptionskip}{4mm} 
		\captionsetup{font={footnotesize,stretch=1.25},justification=raggedleft}
		\caption{\fontsize{10.5bp}{17bp}{Comparison of Scene Segmentation Module}}
		\fontsize{14pt}{14pt}\selectfont 
		\scalebox{0.65}{ 
			\renewcommand{\arraystretch}{1.05}
			\begin{tabular}{c|c|c|c|c|c|c|c|c}
				\Xhline{2pt}
				\makebox[0.045\textwidth]{\multirow{2}{*}{Name}} 
				&\makebox[0.04\textwidth]{\multirow{2}{*}{Size}}
				&\makebox[0.04\textwidth]{\multirow{2}{*}{Out}}
				&\makebox[0.08\textwidth]{\multirow{2}{*}  {$AP_{all}^{@50:95}$}}
				&\makebox[0.055\textwidth] {\multirow{2}{*} {$AP_{all}^{@50}$}} 
				&\makebox[0.055\textwidth]{\multirow{2}{*} {$AP_{all}^{@75}$}}
				& \makebox[0.055\textwidth]{\multirow{2}{*} {$AP_S^{@50}$}} 
				& \makebox[0.055\textwidth]{\multirow{2}{*} {$AP_M^{@50}$}} 
				& \makebox[0.055\textwidth]{\multirow{2}{*} {$AP_L^{@50}$}} \\
				
				 & & & & & & & & \\
				\Xhline{1pt}
				
				M1 & 1 &1 & 44.2 & \uline{94.0} & 35.4 & \uline{93.2} & \uline{96.4} & 89.9 \\
				M2 & 3 & 1 & \uline{44.4} & 93.6 & \textbf{36.5} & 92.9 & 96.2 & \uline{92.0} \\
				M3 & 2&2 & 42.2 & 92.7 & 31.0 & 92.1 & 96.0 & 90.8 \\
				M4 &2 &1 & \textbf{44.5} & \textbf{94.5} & \uline{34.6} & \textbf{93.6} & \textbf{97.3} & \textbf{92.2} \\
\Xhline{2pt}
\end{tabular}}
\vspace{2pt}
\noindent
\setstretch{.8}
{\small \\
Mi (i=1 to 4) represents different experimental groups that use varied methods for generating positive and negative samples to train the Scene Segmentation Module. The “Size" column describes the approach to converting bounding box labels into positive samples for scene segmentation: “1" indicates no expansion, “2" means doubling the size, and “3" means tripling the size. The “Out" column specifies whether clouds and land are differentiated in negative example generation: “1" treats them as one category, and “2" separates them for training.}
\label{Quadruplehead}

\end{table}

\begin{table*}[h!]
\centering
\setlength{\belowcaptionskip}{4mm} 
\captionsetup{font={footnotesize,stretch=1.25},justification=raggedleft}
\caption{\fontsize{10.5bp}{17bp}{Comparison of Data Augmentation and Fine-tuning Strategies}}
\fontsize{14pt}{14pt}\selectfont 
\scalebox{0.65}{ 
\renewcommand{\arraystretch}{1.05}
 \begin{tabular}{c|c|c|c|c|c|c|c|c|c|c|c|c}
 \Xhline{2pt}
 \makebox[0.10\textwidth]{\multirow{2}{*}{Name}} 
 &\multicolumn{4}{c|}{Data Augmentation} 
 &\multicolumn{2}{c|}{Fine-tuning } 
 &\makebox[0.12\textwidth]{\multirow{2}{*} {$AP_{all}^{@50:95}$}}
 &\makebox[0.10\textwidth] {\multirow{2}{*}{$AP_{all}^{@50}$}} 
 &\makebox[0.10\textwidth]{\multirow{2}{*}  {$AP_{all}^{@75}$}}
 & \makebox[0.10\textwidth]{\multirow{2}{*} {$AP_S^{@50}$}}
 & \makebox[0.10\textwidth]{\multirow{2}{*} {$AP_M^{@50}$}} 
 & \makebox[0.10\textwidth]{\multirow{2}{*} {$AP_L^{@50}$}} \\
 
 \Xcline{2-5}{0.4pt}
 & \makebox[0.08\textwidth]{Mixup } 
 & \makebox[0.08\textwidth]{Mosaic} 
 & \makebox[0.08\textwidth]{Affine} 
 &\makebox[0.08\textwidth]{Flip} &\multicolumn{2}{c|}{Strategy} & & & & & & \\
 \Xhline{1pt}
 
 B1 & - & - & - & - &\multicolumn{2}{c|}{-} & 39.0 & 88.0 & 26.8 & 84.7 & 93.9 & 91.8 \\
 B2 & \checkmark & - & - & - &\multicolumn{2}{c|}{Soft} & 40.7 & 93.2 & 28.1 & 92.0 & 96.6 & 90.8 \\
 B3 & - & \checkmark & - & -&\multicolumn{2}{c|}{Soft} & 41.5 & 92.0 & 30.3 & 90.5 & 97.1 & 91.1 \\
 B4 & - & - & \checkmark & -&\multicolumn{2}{c|}{Soft} & 43.1 & 92.5 & \uline{34.7} & 91.8 & \uline{97.3} & 89.4 \\
 B5 & \checkmark& \checkmark & - & - &\multicolumn{2}{c|}{Soft} & 43.7 & 93.4 & 35.4 & 92.5 & 97.1 & \uline{93.1} \\
 B6 &- & \checkmark & \checkmark & - &\multicolumn{2}{c|}{Soft} & 44.1 & 93.2 & 34.6 & 92.0 & 97.2 & 92.9 \\
 B7 & \checkmark & \checkmark & \checkmark & \checkmark&\multicolumn{2}{c|}{Soft} & 42.9 & \uline{93.8} & 31.7 & \uline{92.8} & \textbf{97.5} & 92.8\\
 B8 & -& - & - & \checkmark&\multicolumn{2}{c|}{Soft} & 38.2 & 90.0 & 23.3 & 87.9 & 94.6 & 91.8 \\
 B9 & \checkmark & \checkmark & \checkmark &- &\multicolumn{2}{c|}{$Strategy^1$} & \textbf{44.6} & 93.1 & 38.1 & 91.3 & 96.8 & \textbf{94.6} \\
 B10 & \checkmark& \checkmark &\checkmark& - &\multicolumn{2}{c|}{$Strategy^2$} & 42.5 & 93.2 & 31.1 & 91.8 & 97.1 & 92.1 \\
 B11 &\checkmark& \checkmark & \checkmark &-&\multicolumn{2}{c|}{Soft} & \uline{44.5} & \textbf{94.5} & \textbf{34.6} & \textbf{93.6} & \uline{97.3} & 92.2 \\
 \Xhline{2pt}
 \end{tabular}}
\label{SoftFine-tuning}

\vspace{2pt}
\noindent
\setstretch{.8}
{\small \\
Bi (i=1 to 11) denotes a series of experimental groups that adopt a variety of data augmentation and fine-tuning strategies. 'Soft' refers to the Soft Fine-tuning approach proposed in the study. Strategy¹ involves the training strategy used in YOLOX and YOLOV8, where data augmentation is applied consistently throughout the training period, except for the last 20 epochs, to fine-tune the model. Strategy² ensures a constant 1:1 ratio between augmented and actual data during training. 
}

\end{table*}

\subsubsection{Effect of Gradient-based Module}
 As shown in Table \ref{comparsion_tablet}, the Gradient-based Module effectively enhances the performance of small targets.
 To illustrate the superiority of the Gradient Module proposed in this research, we compared it with two other modules for enhancing small target detection using gradients. One is the convolution kernel with variable weight (FENetwVW) proposed by Hou et al. \cite{17}, which conducts deep feature extraction on top of manual features and effectively detects small targets of different sizes. The other is the multiscale local contrast learning module (MLCL-Net) proposed by Liu et al. \cite{16}, which also performs well in infrared small target detection. The experimental group D1 uses the gradient module FENetwVW, D2 uses the gradient module MLCL, and D5 uses the gradient processing module proposed in this research. These modules demonstrate higher $AP_S^{@50}$ than Net12 in Table \ref{comparsion_tablet}, which does not use gradients, as shown in Table \ref{comparsion_tablet}. By comparing the results of D1, D2, and D5 in Table 5, the experimental results of D5 are 1.3\% and 1.5\% higher in $AP_{all}^{@50}$ compared to D1 and D2, respectively.
To demonstrate the effectiveness of both first-order and second-order transformations, we conducted experiments D3 and D4, respectively, with the results shown in Table \ref{Dual}. We use gradients in D3, which results in a 1.5\% improvement in $AP_S$ compared to Net12 in Table \ref{comparsion_tablet}. When comparing D4 with Net12, we find that using the second-order transformation alone does not significantly enhance the network. However, the use of both first-order and second-order gradients can improve network performance, with $AP_{all}^{@50}$ increasing by 1.1\%, indicating a significant improvement.

\begin{table}[h!]
 \centering
 \setlength{\belowcaptionskip}{4mm} 
 \captionsetup{font={footnotesize,stretch=1.25},justification=raggedleft}
 \caption{\fontsize{10.5bp}{17bp}{Performance Comparison of The State-of-the-art Networks on IRSDSS}}
 \fontsize{14pt}{14pt}\selectfont 
 \scalebox{0.65}{ 
 \renewcommand{\arraystretch}{1.05}
 \begin{tabular}{c|c|c|c|c|c|c|c}
 \Xhline{2pt}
 \makebox[0.06\textwidth]{\multirow{2}{*}{Name}} 
 &\makebox[0.08\textwidth]{\multirow{2}{*} {$AP_{all}^{@50:95}$}}
 &\makebox[0.055\textwidth]{\multirow{2}{*} {$AP_{all}^{@50}$}} 
 &\makebox[0.055\textwidth]{\multirow{2}{*} {$AP_{all}^{@75}$}}
 &\makebox[0.055\textwidth]{\multirow{2}{*} {$AP_S^{@50}$}} 
 &\makebox[0.055\textwidth]{\multirow{2}{*} {$AP_M^{@50}$}} 
 &\makebox[0.055\textwidth]{\multirow{2}{*} {$AP_L^{@50}$}} 
 &\makebox[0.03\textwidth]{\multirow{2}{*} {Size}} \\
 & & & & & & & \\
 \Xhline{1pt}
 Faster RCNN &38.0 &87.4&24.2&86.0&93.4&82.6&165.0\\
 RetinaNet &36.9&87.3&21.6&84.3&95.0&88.3&245.4\\
 SSD & 34.8 & 88.6 & 17.1 & 88.6 & 93.4 & 91.9 &186.1 \\
 YOLOv5 & 42.6 & 92.7 & 30.5 & 92.4 & 97.2 & 88.7 &88.5 \\
 YOLOv8 &42.5 &91.8 & 31.3 & 90.8 & 95.3 & 90.3 &\textbf{49.6} \\
 RT-DETR &42.2 &91.4 &29.9 &91.4&93.0&87.3&63.0\\
 KCPNet & 43.9 & 93.4 & 33.9 & 92.5 & 97.2 &93.1 &731.2\\
 ours &\textbf{44.5} & \textbf{94.5} & \textbf{34.6} & \textbf{93.6}&\textbf{97.3} & \textbf{92.2} & 53.8\\
	\Xhline{2pt}
	\end{tabular}}
	\label{state_of_art}
\end{table}

\subsubsection{Effect of Scene Segmentation Module}
This research adds an environment-aware head to the network to perform scene semantic perception. The introduction of the mask module (Net4) raises $AP_{all}^{@50}$ to 91.2\% and boosts $AP_L^{@50}$ by 1.5\%, compared to the baseline network in Table \ref{comparsion_tablet}. Furthermore, the positive samples required for training are generally directly converted using bounding box labels. In Table \ref{Quadruplehead}, M1 adopts the original bounding box sizes to generate masks, while M2 and M4 correspond to enlarging the bounding boxes by factors of 3 and 2, respectively. Maintaining the original box size and enlarging the target by a factor of three do not provide a sufficient improvement in performance compared to enlarging the target by a factor of two. We also explored the generation rules for negative samples. As presented in Table \ref{Quadruplehead}, the M4 group, by amalgamating land and clouds into a unified background category, achieves a superior enhancement in network performance relative to the M3 group, wherein land and clouds are considered independent classification labels during the training phase. Thus, it is evident that considering convective clouds and land separately as two different negative examples is ineffective.

\subsubsection{Effect of Soft Fine-tuning}
We design the training strategy, Soft Fine-tuning, to solve the distortion caused by strong data augmentation. By incorporating the Soft Fine-tuning strategy into the network training process, we successfully elevated $AP_{all}^{@50}$ from 90.9\% to 92.1\%, as depicted in Table \ref{comparsion_tablet}. Further integrating the modules proposed in this research with the Soft Fine-tuning training strategy improves performance metrics. 
To illustrate, comparing Net3 with Net8 in Table \ref{comparsion_tablet}, $AP_{all}^{@50:95}$ improved from 40.9\% to 43.3\%, $AP_{all}^{@50}$ from 91.6\% to 93.0\%, and $AP_{all}^{@75}$ from 29.2\% to 34.2\%. 
Moreover, the $AP_{all}^{@50}$ across various networks using Soft Fine-tuning in Table \ref{comparsion_tablet} also exhibited growth from 0.2\% to 2.6\%. 
Overall, Soft Fine-tuning demonstrates excellent performance as an innovative training strategy, enhancing AP metrics across networks when combined with various modules. In Table \ref{SoftFine-tuning}, we further explore the impact of Soft Fine-tuning and data augmentation strategies on network detection performance. Table \ref{SoftFine-tuning} compares the impact of different data augmentations on the detection performance of the network. Comparing B1, B2, B3, and B4 shows that Mixup, Mosaic, and affine transformation can improve APs. 
For example, the affine transformation can improve the $AP_{all}^{@50:95}$, which is as high as 4.1\%. Mixup and Mosaic can increase the $AP_{all}^{@50}$ value by 5.2\% and 4.0\%, respectively. 
It is worth noting that in the B7 and B11 test groups, the AP value does not increase but decreases with the flip, which shows that flipping is inappropriate for ship detection. 
To compare the effects of different training strategies on network performance, we experimented with three distinct approaches for training with augmented data. The B9, B10, and B11 test groups in Table \ref{SoftFine-tuning} serve as comparisons. Group B9 utilized the training strategy, employing augmented data throughout the training process and fine-tuning with original data in the final epochs. Group B10 followed the training strategy, maintaining a 1:1 ratio of augmented to original data and ensuring continuous training. Group B11 applied the Soft Fine-tuning training strategy proposed in this study, using both augmented and original images during training with a gradually decreasing proportion of augmented images. The Soft Fine-tuning proposed in this research has apparent advantages over the other two training strategies.

\subsection{Comparison With the State-of-the-Art (SOTA)}

\begin{figure}
 \centering
 \includegraphics[scale=.25]{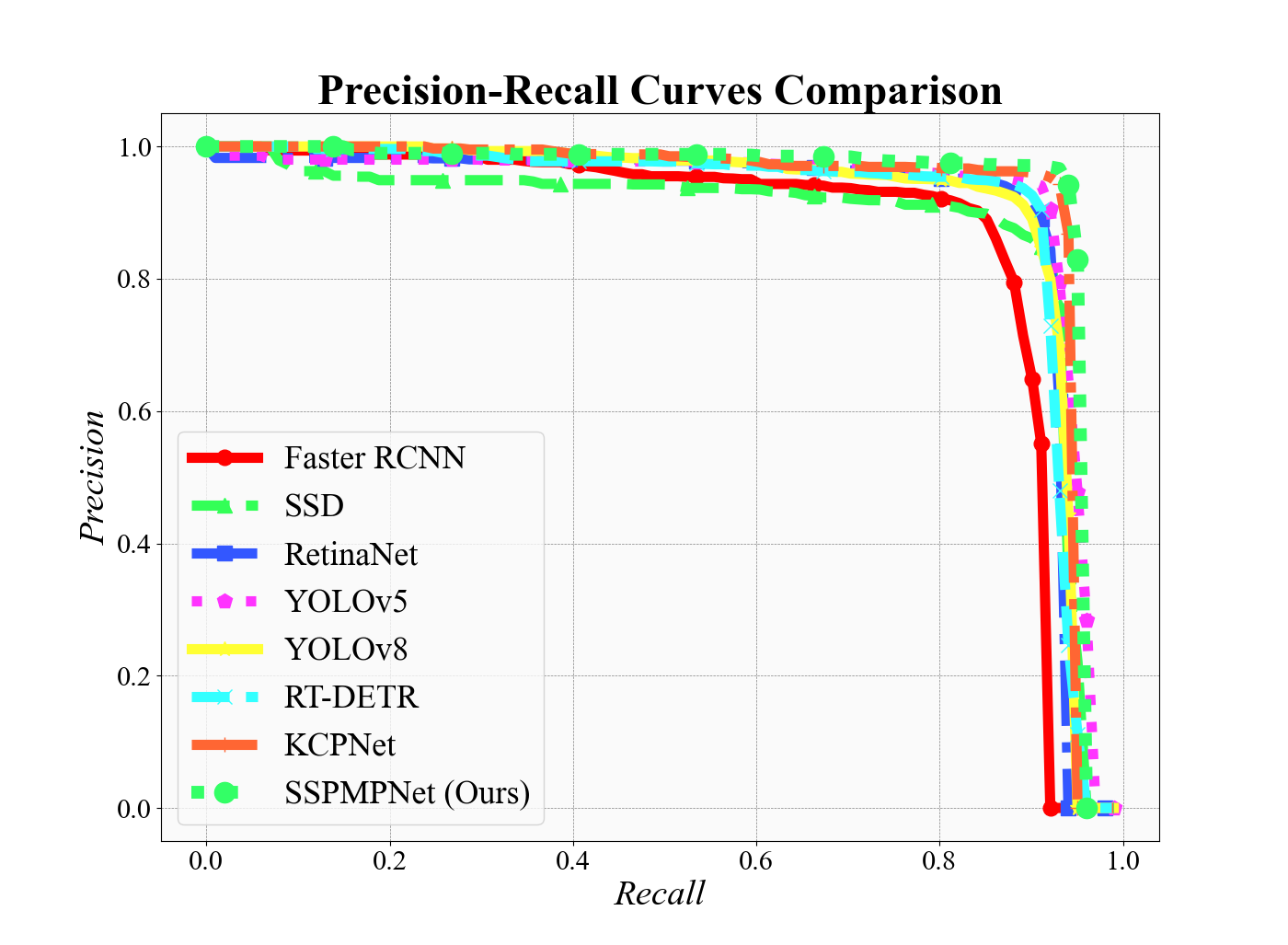}
 \caption{The precision-recall curves of various networks.}
 \label{fig_22}
\end{figure}

To evaluate the performance of our proposed approach, we compare our network with well-established networks, including SOTAs, using the IRSDSS dataset. The P-R curve is shown in Fig. \ref{fig_22}.

We benchmark our network against a classic network. 
The results indicate that our network achieves substantial improvements over traditional one-stage models, RetinaNet and SSD, with performance increases of up to 7.6\% in $AP_{all}^{@50}$, 13\% in $AP_{all}^{@75}$, and 9.3\% in $AP_S^{@50}$. We also compare our model with the conventional two-stage detection networks, like Faster RCNN. Not only is our model significantly more compact, but it also consistently delivers higher APs across all evaluated categories.

We compared our network with the latest and widely used object detection methods. We compare our network with the YOLO series, widely used in target detection due to its fast speed, high precision, and compact model. As depicted in Table \ref{state_of_art}, our network outperforms the leading YOLOv5 and YOLOv8, evidencing a substantial enhancement with a 1.9\% increase in $AP_{all}^{@50:95}$, a 1.8\% improvement in $AP_{all}^{@50}$ over YOLOv5, and a notable 3.3\% rise in the $AP_{all}^{@75}$ compared to YOLOv8. We also compare our network with the latest transformer-based detection networks, RT-DETR. Our method demonstrates improved performance over RT-DETR \cite{57}, with higher precision gains ranging from 2.2\% to 4.9\% across various APs. It is also more storage-efficient, being 9.2 MB smaller.

Furthermore, we compare our method with the current optimal IRSD methods. Our network is smaller and more precise than KCPNet \cite{13}, offering better performance in infrared image detection and classification. 

\subsection{Disscusion}
In the Gradient-based Module, the weight of each direction varies for various targets. However, the importance of the eight directions is different for various targets. We are supposed to give the eight directions various weights for diverse targets. Therefore, the Gradient-based Model can be further improved.

For the Scene Segmentation Module, the module based on transformers should be introduced to the network because transformers could incorporate and fully exploit ships' long-distance dependency, which is beneficial for scene perception. 

In practical applications, target classification is essential in addition to IRSD. The slight differences between infrared ships necessitate the development of a more sophisticated and powerful neural network.

\section{Conclusion}

To address the challenges of IRSD in complex backgrounds, we propose SMPISD-MTPNet, a new network for ship detection. The research's primary contributions include SSE to extract scene semantic prior and guide the network, the Multi-task Perception Module, and the Soft Fine-tuning. We introduce SSE, which leverages the characteristics of infrared ships and prior knowledge to enrich the semantics of targets. 
We present the Multi-Task Perception Module, which incorporates scene segmentation as an auxiliary task and utilizes the Gradient-based Module with specialized gradients to enhance the detection of small and dim infrared ships.
Then, we introduce a novel training approach termed Soft Fine-tuning, designed to mitigate the distortions introduced by data augmentation. To improve the detection capabilities of our scheme, we plan to embed the infrared radiation characteristics of various targets and backgrounds in deep learning framework, thereby enhancing the accuracy of IRSD. Furthermore, we will expand our exploration mission from detection to recognition for infrared ship in the future.

\vfill
\end{document}